\documentclass[journal]{IEEEtran}
\usepackage{amsmath,amsfonts}
\usepackage{algorithmic}
\usepackage{algorithm}
\usepackage{array}
\usepackage[caption=false,font=normalsize,labelfont=sf,textfont=sf]{subfig}
\usepackage{textcomp}
\usepackage{stfloats}
\usepackage{url}
\usepackage{verbatim}
\usepackage{graphicx}
\usepackage{cite}
\usepackage{hyperref}
\usepackage{url}
\usepackage{bm}
\newcommand*{\mc}[1]{\mathcal{#1}}

\renewcommand*{\b}[1]{\bm{#1}}

\usepackage{dsfont}
\newcommand*{\ds}[1]{\mathds{#1}}

\newcommand*{\wh}[1]{\widehat{#1}}
\newcommand*{\wt}[1]{\widetilde{#1}}

\let\span\relax
\DeclareMathOperator{\span}{span}

\DeclareMathOperator{\diag}{diag}

\DeclareMathOperator{\argmin}{arg\,min}

\newcommand{\Gr}{\mathrm{Gr}(n,k)}
\newcommand{\St}{\mathrm{St}(n,k)}

\newcommand{\bV}{V_\bullet}
\newcommand{\bD}{D_\bullet}
\usepackage{xcolor}
\input{rwth-color.sty}
\usepackage{enumitem}
\usepackage{amssymb}

\usepackage{tikz}
\usepackage{tikz-network}

\newtheorem{theorem}{ Theorem}[section]
\newtheorem{definition}{ Definition}
\newtheorem{remark}{ Remark}[section]
\newtheorem{problem}{Problem}
\newtheorem{lemma}{Lemma}[section]
\newtheorem{proposition}{ Proposition}

\begin{document}

\title{Grassmanian Interpolation of Low-Pass Graph Filters: \\ Theory and Applications}

\author{Anton Savostianov, Michael T. Schaub, Benjamin Stamm
	% <-this % stops a space
	\thanks{A.~Savostianov and M.T.~Schaub are with Computational Network Science, RWTH Aachen, Germany.  M.T.Schaub is with Center for Computational Life Sciences, RWTH Aachen, Germany. B.~Stamm is with University of Stuttgart, e-mails: \url{savostianov@cs.rwth-aachen.de}, \url{schaub@cs.rwth-aachen.de}, \url{benjamin.stamm@ians.uni-stuttgart.de}}% <-this % stops a space
	\thanks{Manuscript received October, 2025.}
}

% The paper headers
\markboth{Transactions on Signal Processing}%
{Savostianov \MakeLowercase{\textit{et al.}}: Interpolation of Low-Pass Graph Filters}

%\IEEEpubid{0000--0000/00\$00.00~\copyright~2021 IEEE}
% Remember, if you use this you must call \IEEEpubidadjcol in the second
% column for its text to clear the IEEEpubid mark.

\maketitle

\begin{abstract}
	Low-pass graph filters are fundamental for signal processing on graphs and other non-Euclidean domains.
	However, the computation of such filters for parametric graph families can be prohibitively expensive as computation of the corresponding low-frequency subspaces, requires the repeated solution of an eigenvalue problem.
	We suggest a novel algorithm of low-pass graph filter interpolation based on Riemannian interpolation in normal coordinates on the Grassmann manifold.
	We derive an error bound estimate for the subspace interpolation and suggest two possible applications for induced parametric graph families.
	First, we argue that the temporal evolution of the node features may be translated to the evolving graph topology via a similarity correction to adjust the homophily degree of the network.
	Second, we suggest a dot product graph family induced by a given static graph which allows to infer improved message passing scheme for node classification facilitated by the filter interpolation. 
\end{abstract}


\section{Introduction.}
\label{sec:intro}

Graph Signal Processing (GSP) generalizes ideas from classical signal processing to non-Euclidean domains without requiring equidistant nodes, regular grids, or an explicit spatial/temporal embedding \cite{ortegaGraphSignalProcessing2018,stankovic2018introduction}.
In GSP, scalar or multidimensional signals \( \b x \) are supported on the vertices of a graph \( \mc G \) and processed via a graph shift operator \( S \).
The graph \( \mc G \) encodes the topology of an abstract relational domain, making the setting highly expressive.
As a result, GSP provides a natural interface for tasks such as reconstruction of missing values, vertex classification, signal smoothing, and clustering \cite{you2020handling,narang2013signal,sandryhailaDiscreteSignalProcessing2013,chen2014signal}, across applications centered on network-structured data (e.g., traffic and social networks, electrical grids, molecular and sensor graphs) \cite{qiu2017time,shuman2013emerging,li2021graph,mateos2019connecting}.
Moreover, the composition of linear message passing with nonlinear activation functions forms the principal structure of graph neural networks (GNNs) \cite{bronstein2017geometric}.
GSP extends naturally to higher-order domains with signals on edges or faces when the underlying relational model is a hypergraph, cell complex, or simplicial complex \cite{schaub2018flow,schaub2022signal,roddenberry2022signal}.

Analogous to the Euclidean setting, vertex signals \( \b x \) are processed via linear operators (filters) \( H \), \( \b x_{\mathrm{out}} = H \b x_{\mathrm{in}} \).
Of particular interest are shift-invariant filters, which satisfy \( SH = HS \) and admit a spectral representation \( S = V \Lambda V^\top \), \( H = V h(\Lambda) V^\top \), where \( h \) is the frequency response.
A common and scalable choice is to take \( h \) polynomial, \( h(\lambda) = \sum_{i=0}^{M-1} h_i \lambda^i \), yielding \( H = \sum_{i=0}^{M-1} h_i S^i \) and realizing localized propagation via powers of \( S \) \cite{ortegaGraphSignalProcessing2018}.
Owing to its spectral properties, the graph Laplacian \( L \) is a typical choice of \( S \) \cite{spielman2012spectral}.
As with the circular shift operator on regular grids, eigenvectors of \( L \) associated with smaller eigenvalues (low frequencies) correspond to smooth signals (low variation across edges), whereas high-frequency eigenvectors capture oscillatory or noisy components.
This motivates low-pass filters that (ideally) project onto the span of low-frequency eigenvectors, retaining structural content while suppressing high-variation noise \cite{ramakrishna2020user,nt2019revisiting}.

Computationally, such low-pass filters hinge on extracting the low-frequency subspace \( \mc V \).
This becomes numerically demanding for time-evolving or parametric graph families \( \mc G(t) \) (e.g., transportation networks, electrical grids, sensor networks), where the principal subspace \( \mc V(t) \) varies with the parameter \( t \).
We propose an efficient interpolation approach for low-pass graph filters over parametric families \( \mc G(t) \).
We note that the trajectory \( \mc V(t) \) falls on the Grassmann manifold, and describes a set of fixed-dimensional subspaces; numerically, points on the manifold can be represented by orthogonal projectors onto the subspace or by equivalence classes of orthonormal bases.
Because the Grassmann manifold is curved, direct application of standard polynomial interpolation (e.g., Lagrange polynomials) may lead to an interpolant outside the manifold.
Hence, we adapt Riemannian interpolation in normal coordinates, i.e., we interpolate in the tangent space of a base point and map back via the exponential, following \cite{zimmermannManifoldInterpolationModel2022,bendokat2024grassmann}.
For this we build on successful applications in parametric eigenvalue problems for electronic structure calculations \cite{polack2020approximation,polack2021grassmann,pes2023quasi}.
We derive a novel error bound for subspace interpolation in normal coordinates (Theorem~\ref{thm:int_err}) showing that interpolation error in the tangent space propagates linearly to the manifold.
A related bound has recently been shown for interpolation in maximum-volume coordinates \cite{jensen2025maximum}.
Leveraging this guarantee, we design an algorithm for Riemannian interpolation of graph filters that incorporates spectrum updates and realignment of the interpolated subspace.

We then demonstrate the approach on two induced parametric graph families. First, because the performance of graph filters and GNNs often depends on the homophily/heterophily level, downweighting heterophilic edges in homophilic networks can improve task performance \cite{ma2021homophily}. When vertex features are time‑dependent, homophily may vary over time; we propose a time‑varying similarity correction of edge weights that downweights heterophilic edges and use the resulting family \( \mc G(t) \) to showcase our interpolation algorithm and its computational savings. Second, for any static graph \( \mc G \), we construct a parametric family of dot product graphs \( \mc G(\delta) \) that resemble the original while enabling additional connections and rewiring \cite{athreya2018statistical,marencoOnlineChangePoint2022}. Concretely, we obtain low‑rank embeddings of vertices in \( \ds R^d \) from a truncated spectral approximation of \( S \) and form edges between pairs whose inner products exceed a threshold \( \delta \), which serves as the family parameter. We argue that suitable members of \( \mc G(\delta) \) preserve the static structure while improving message passing; searching for the optimal \( \delta \) at scale becomes tractable via our filter interpolation framework. Finally, we show on a simple vertex classification task via graph filters \cite{sandryhailaDiscreteSignalProcessing2013} that one can indeed find \( \mc G(\delta) \) achieving improved classification accuracy.

\subsection*{Outline}
Section~\ref{sec:filters} introduces the requisite background on graphs and low‑pass filters. Section~\ref{sec:intrepolation} formulates filter interpolation for parametric graph families, presents Grassmannian interpolation in normal coordinates for eigenspaces, and details our interpolation scheme for graph filters. The error bound is derived in Subsection~\ref{subsec:error}. Sections~\ref{sec:csbm} and \ref{sec:dpg} apply the method to (i) a graph family induced by a time‑varying similarity correction for vertex features and (ii) a dot product graph family that facilitates improved message passing for vertex classification. Section~\ref{sec:conclusion} concludes and outlines future directions.

\section{Graph Filters}
\label{sec:filters}

\subsection{Graphs and matrix representations.}
A (weighted) \emph{graph} \( \mc G \) is defined as a triplet \( \mc G = \left\{ \ds V, \mc E, w(\cdot) \right\} \) where \( \ds V \) is a set of vertices, \( \mc E \subseteq \ds V \times \ds V \) is a set of edges, and \( w: \mc E \mapsto \ds R_+ \) is a weight function; let \( | \ds V | = n \) and \( | \mc E | = m \).
Additionally, we assume that \( \mc G \) is the undirected graph, so if \( [v_i, v_j] \in \mc E \Rightarrow [v_j, v_i] \in \mc E \).
A graph shift operator \( S \in \ds R^{n \times n}\) is a linear operator encoding the sparsity pattern of the graph structure, whose action describes the propagation of information from one vertex to its neighbours.
Formally, we have that \( S_{ij} = 0 \) if \( [v_i, v_j] \notin \mc E \) and \( i \ne j \); the diagonal entries \( S_{ii} \) can be arbitrary.
Common examples of shift operators include the adjacency matrix \( A \), defined by its entries \( A_{ij} = w([v_i, v_j]) \) if \( [v_i, v_j] \in \mc E \) and the Laplacian matrix \( L \), which is defined via \( L_{ij} = -w([v_i, v_j]) \) if \( [v_i, v_j] \in \mc E \) and \( L_{ii} = \sum_{[v_i, v_j] \in \mc E} w([v_i, v_j]) \).
We use the matrix \( D = \diag \left( A \b 1 \right) \) to denote the diagonal matrix of vertex (weighted) degrees.
We can then write \( L = D - A \). Other alternatives for the shift operator include the normalized Laplacian \( \tilde L = D^{-1/2} L D^{-1/2} \) and the random walk Laplacian \( \hat L = I - D^{-1} A \).

\subsection{Filters.}

Graph signal processing (GSP) considers signals on the vertices of a graph.
We will represent a signals on vertex $i$ by a vector \( \b x_i \in \ds R^d \); for simplicity, below we focus on the case of the scalar signals, \( d = 1 \).
The collection of the signals for all vertices is referred to as \( \b x \in \ds R^n \) (in the case of multidimensional signals, one collects vertex signals into a matrix \( X \in \ds R^{n \times d}\)).
Common GSP tasks include vertex classification, reconstruction of missing values, de-noising or segmentation of the signal, and so on.
These tasks are typically processed via linear operators \( H \) known as filters, i.e. \( \b x_{out} = H \b x_{in}\).
Frequently, one is interested in shift-invariant graph filters, which commute with the graph shift operator such that \( H S = S H\).
The action of such a shift-invariant filter can be represented in the form of the frequency response of the filter, \( H = V h( \Lambda ) V^\top \), where \( S = V \Lambda V^\top \) is the spectral decomposition of the shift operator \( S \) and \(h(\Lambda ) \) is an arbitrary function.
Since high-magnitude eigenvalues of \( S \) are associated with rapidly changing signals~\cite{ortegaGraphSignalProcessing2018}, which are often associated to noise in the associated signals, in our work we focus on low-pass filters.

Note that an arbitrary frequency response \( h(\Lambda) \) can be approximated by a polynomial \( p_M (\Lambda) \) of degree \( M-1 \), so the resulting filter \( H \) of a graph shift operator \( S = V \Lambda V^\top \) is given by
\begin{equation*}
	H = V p_M(\Lambda) V^\top = V \left( \sum_{i=0}^{M-1} h_i \Lambda^i \right) V^\top = \sum_{i=0}^{M-1} h_i S^i
\end{equation*}
where filter coefficients \( h_i \) are known as filter \emph{taps}.
Let \( \Psi \) be a Vandermond matrix of the spectrum \( \sigma( S ) = \{ \lambda_1, \lambda_2, \dots, \lambda_n \} \) defined as \( \Psi_{ij} = \lambda_i^{j-1}\); then
\begin{equation*}
	H = V \diag \left( \Psi \b h \right) V^\top
\end{equation*}
with \( \b h \) being the vector of filter taps.

\begin{definition}[Low-Pass Graph Filter]
	Graph filters which only let signals pass that are composable from eigenvectors with small frequencies (eigenvalues) are known as \emph{low-pass filters}.
	Specifically, a \( k \)-lowpass filter is defined as:
	\begin{equation*}
		H^{(k)} = V^{(k)} \diag( \Psi^{(k)} \b h ) ( V^{(k)} )^\top,
	\end{equation*}
	where \( V^{(k)} \in \ds R^{n \times k }\) is a matrix of the first \( k \) unit eigenvectors (corresponding to the smallest eigenvalues) and \( \Psi^{(k)}\) is the corresponding minor of \( \Psi \) of the first \(k\) rows.
	If \( k \) is fixed and specified, we omit the index for brevity.
\end{definition}

Naturally, the low-pass filter \( H \) should not depend on the choice of the basis in the eigenspaces.
Note that \( H \) can be reformulated as the sum of spectral projectors weighted by the corresponding frequency responses, \( H = \sum_{i=1}^k p_M( \lambda_i ) \b v_i \b v_i^\top \).
It is thus sufficient to require that \( V^{(k)} \) does not to break any eigenspaces, such that \( \{ \lambda_1, \ldots \lambda_k \} \cap \{ \lambda_{k+1} \ldots \lambda_n \} = \varnothing\), as the sum of the spectral projectors is invariant under orthogonal transformations of eigenvectors corresponding to the same eigenvalue.
As a result, each low-pass filter is supported by the direct sum of the eigenspaces of the shift operator.

\section{ Eigenspace interpolation }
\label{sec:intrepolation}

\subsection{Parametric graph families and graph filters}

The concept of a low-pass graph filter extends naturally to parametric graph families.
Consider a trajectory of graphs \( \mc G(t) = (\mc V, \mc E_t, w_t) \) over a real parameter \( t \in \ds R \) (often interpreted as time), with a fixed vertex set and time-varying edges/weights. Let \( S(t) \) denote the shift operator at time \( t \) (e.g., the unnormalized or normalized Laplacian \( L(t) \), or the adjacency \( A(t) \)).
In this section we primarily take \( S(t)=L(t) \) for concreteness; the development is otherwise agnostic.

Straightforward examples of such families include networks induced by discretizations of deforming bodies \cite{pilvaLearningTimedependentPDE2022} or filtrations such as Vietoris–Rips complexes often used in topological data analysis \cite{hausmann1995vietoris,edelsbrunner2010computational,chazal2021introduction}. The corresponding low-pass filter, supported on the first \( k \) eigenmodes of \( S(t) \), can be written as
\[
	H(t) = V(t)\,\mathrm{diag}\big(\Psi(t)\,\b h(t)\big)\,V(t)^\top,
\]
where \( S(t) V(t) = V(t)\Lambda(t) \), \( \Lambda(t) = \mathrm{diag}(\lambda_1(t),\dots,\lambda_k(t)) \), and \( \Psi(t) \) is the Vandermonde matrix of size \( k \times M \) associated with the polynomial frequency response \( h(\lambda) = \sum_{i=0}^{M-1} h_i \lambda^i \), i.e.,
\[
	\Psi(t) = \begin{bmatrix}
		1      & \lambda_1(t) & \cdots & \lambda_1(t)^{M-1} \\
		\vdots & \vdots       &        & \vdots             \\
		1      & \lambda_k(t) & \cdots & \lambda_k(t)^{M-1}
	\end{bmatrix},\quad
	\b h(t) = \begin{bmatrix}h_0\\\vdots\\h_{M-1} \end{bmatrix}.
\]
Equivalently, we may write \( H(t) = \sum_{i=0}^{M-1} h_i\,S(t)^i \), linking spectral filtering to localized propagation via powers of \( S(t) \).

\begin{remark}[From eigenvectors to subspaces]
	When an eigenvalue has multiplicity greater than one, any orthonormal basis of its eigenspace is defined only up to an orthogonal rotation. Consequently, eigenvector trajectories may be discontinuous even when the underlying subspace trajectory is perfectly smooth.
	It is therefore more natural to work with the evolving subspace \( \mc V(t) = \mathrm{im}\,V(t) \) (a point on the Grassmann manifold) rather than a particular eigenvector basis.
\end{remark}

\begin{remark}
	The Vandermonde matrix \( \Psi(t) \) depends only on the spectrum \( \Lambda(t) \), which can be computed as \( \Lambda(t) = V(t)^\top S(t) V(t) \). This underscores that the dominant computational cost in the parametric setting is obtaining the low-frequency subspace \(\mathrm{im} V(t) \).
\end{remark}

\subsection{Grassmann manifold and interpolation}

We focus on the task of recovering the extremal (low-frequency) subspace \( \mc V(t) \) and the corresponding spectrum \( \Lambda(t) \) at arbitrary \( t \), given exact evaluations of these  objects at certain ("anchor") parameters.
As direct recomputation of \( \mc V(t) \) at each point \( t \) is often prohibitive, we propose interpolation on the Grassmann manifold.

\begin{problem}
Assume the exact values of \( \mc V(t) \) along a graph trajectory are available at \( N \) moments in time, \( t_1, t_2, \dots t_N \), with \( V_1 = V(t_1), \dots V_N = V( t_N )\) (with fixed dimension of the subspace \( k \)) be corresponding bases; find the interpolation \( \wt V(t) \) for the basis \( V(t) \) for arbitrary \( t \).
\end{problem}

This problem cannot be tackled using classical interpolating approaches due to the following two reasons: classical interpolation methods such as Lagrange interpolation do not preserve the orthonormal structure of the matrix \( V(t) \); more importantly, we are interested in interpolating subspaces \( \mc V(t)\) spanned by eigenvectors, not the specific eigenvectors themselves. Our interpolation routine should thus be rotation-invariant, which is not the case for classical interpolation methods.
These issues stem from the fact that by its definition, the subspace trajectory \( \mc V(t) \) lies on the Grassmann manifold \( \Gr \) of \( k \)-dimensional subspaces of \( \ds R^n \), which is a non-linear space.

\begin{definition}[Grassmann manifold]
	The Grassmann manifold (Grassmannian) \( \Gr \) is the set of all \( k \)-dimensional subspaces of \( \mathbb{R}^n \):
	\begin{equation*}
		\Gr = \left\{  \mc V \subset \ds R^n \middle| \mc V \text{ is a subspace and } \mathrm{dim}\,\mc V = k \right\}
	\end{equation*}
\end{definition}
By its definition, the subspace trajectory \( \mc V (t) \) induced by \( \mc G(t) \) is a curve on the Grassmann manifold \( \Gr \).

Computationally, it is instrumental to work with a matrix-based representation of the Grassmann manifold as the manifold of linear subspaces.
To this end, we can leverage the following ideas.

First, we represent a subspace \( \mc V \) via its orthonormal basis \( V = ( \b v_1 \mid \b v_2 \mid ... \mid \b v_k )\).
Note that, \( V^\top V = I \) and the set of such orthogonal matrices \( V \) of rank \( k \) is known as \textit{Stiefel manifold}.
However, a \emph{Stiefel representative} \( V \) for the subspace \( \mc V \) is not unique up to orthogonal transformation \( Q \in \ds R^{k \times k}\) with \( Q^\top Q = Q Q^\top = I \).
As a result, each element of the Grassmann manifold \( \Gr \) is isomorphic to the equivalence class \( [ V ]\):
\begin{equation*}
	\begin{aligned}
		\mc V \sim [V] = \left\{ V Q \quad \text{for any orthogonal } Q \in \mathbb{R}^{k \times k}, \right. \\ \left. Q Q^\top = Q^\top Q = I \right\}.
	\end{aligned}
\end{equation*}
Stated differently, the Grassmann manifold \( \Gr \) is a quotient space \( \St / \mathbb{O}(k) \) where \( O(k) \) is the orthogonal group of \( k \times k \) matrices.

While the mapping to the equivalence class is natural, it is not a matrix representation per se.
Instead, we relate the subspace \( \mc V \) to the corresponding orthogonal projector \( D \).
Indeed, note that for any Stiefel representative \( V \in [V] \cong \mc V \), the corresponding projector \( D \) is given by \( D = V V^\top \) which is the same for any \( V \in [V] \).
Hence, the Grassmannian \( \Gr \) can be identified with the set of all \( n \times n \) orthogonal projectors \(  D \) of rank \( k \):
\begin{equation*}
	\begin{aligned}
		\Gr \cong \{ D \in \ds R^{n \times n} \mid D^\top = D, \; D^2 = D, \\ \mathrm{rank}(D) = k\}
	\end{aligned}
\end{equation*}
To facilitate computations, each projector \( D \) is still stored via some corresponding \( V \in [V] \) such that \( D = V V^\top \).
We refer to any matrix \( D \in \Gr \) as a Grassmann matrix and to a corresponding Stiefel representative \( V \).
Following these considerations, each graph trajectory \( \mc G(t)\) corresponds to a subspace trajectory \( \mc V(t) \in \Gr \) with corresponding equivalence class trajectory \( [V](t) \), projector trajectory \( D(t)\) and multiple trajectories Stiefel representatives \( V(t) \). we use all these characterizations interchangeably.

Using the now introduced notation, we can reformulate the problem of interpolating eigenspaces as follows:
\begin{problem}
For a fixed dimension \(k\) of the eigenspace, assume that the exact computation of \( \mc V(t) \) along the graph trajectory \( \mc G_t \) is available at \( N \) moments in time, \( t_1, t_2, \dots t_N \), with Stiefel representatives \( V_1 = V(t_1), \dots, V_N = V( t_N ) \) and corresponding Grassmann matrices \( D_1 = V_1 V_1^\top, \dots D_N = V_N V_N^\top \).
Find the interpolation \( \wt D(t) \) of the projector \( D(t) \in \Gr \) (and its Stiefel representative \( \wt V(t)\)) for arbitrary \(t\).
\end{problem}

\subsection{Interpolation in normal coordinates}
The task of interpolating on a non-trivial manifold can be reduced if we can establish an intermediate vector space where common interpolation methods such as Lagrangian interpolation can be applied.
A typical approach in this direction is the Riemannian interpolation method in the normal coordinates which we describe next (see \cite{zimmermannManifoldInterpolationModel2022, bendokat2024grassmann}, or \cite{ciaramella2025gentle} for more pedagogical introduction).

We first provide a brief overview of the notions required for the interpolation in normal coordinates on the Grassmann manifold \( \Gr \), to gain some high-level intuition.
The formal definitions are provided thereafter.
Namely, let \( \bD \in \Gr \) (or \( [ \bV ] \in \Gr \)) be a fixed base point and consider the tangent space \( T_{ \bD } \Gr \) corresponding to this base point.
The key idea of the interpolation in the normal coordinates is to map each Grassmanian \( D_i\) to a point $\Delta_i$ on the tangent space of the base point \( T_{ \bD } \Gr \), interpolate between all those points in the tangent space, and then map back the interpolant from  the tangent space \( T_{ \bD } \Gr \) to the manifold \(\Gr \).

\begin{figure}[!t]
	\centering
	\scalebox{0.35}{\begin{tikzpicture}

			\draw[line width = 3px, rwth-black-50] (-1, -10) -- (2, -4.5) -- (13, -4.5) -- (10, -10) ;
			\node[scale = 3.0] at (13.5, -10) { \color{rwth-black-50} \( T_{\bD} \Gr \) };

			\draw[line width = 5px,] plot [smooth, tension=1] coordinates { (0,0) (2,2) (4,-4) (8,0) (10,2) (12,0) (16,4) (16,0) (18,-2) (20,0)};

			\draw[line width=5px, dashed] plot [smooth, tension=1] coordinates { (4, -9) (14, -7) (18, -2)};
			\draw[line width=5px, dashed] plot [smooth, tension=1] coordinates { (4, -9) (12, -4) (16, 4)};
			\draw[line width=5px, dashed] plot [smooth, tension=1] coordinates { (4, -9) (7, -3) (8, 0)};
			\draw[line width=5px, dashed] plot [smooth, tension=1] coordinates { (4, -9) (-2, -3) (2, 0) (2, 2)};
			\draw[line width=6px, dashed, color = rwth-blue] plot [smooth, tension=1] coordinates { (4, -9) (10, -4) (12,0)};

			\draw[-latex, line width = 5px] (4, -9) -- (1, -7);
			\node[scale = 2.0, fill=rwth-black-25] at (-0.5, -8.6){ \( \mathrm{Log}_{[\bV]}^{\Gr}(V_1) =  \Delta_1\)};
			\draw[-latex, line width = 5px] (4, -9) -- (6.5, -5);
			\node[scale = 2.5] at (4.3, -6.5){ \( \Delta_2\)};
			\draw[-latex, line width = 5px] (4, -9) -- (8, -7.7);
			\node[scale = 2.5] at (8.5, -8.0){ \( \Delta_3\)};
			\draw[-latex, line width = 5px] (4, -9) -- (10, -8.9);
			\node[scale = 2.5] at (8, -9.6){ \( \Delta_N\)};
			\draw[-latex, line width = 5px, color = rwth-blue] (4, -9) -- (9, -5.5);
			\node[scale = 2.0, fill=rwth-blue-25] at (12.1, -5.3){\large \textcolor{rwth-blue}{\( \wt\Delta = \sum \ell_j (\wt t) \Delta_j \)}};

			\Vertex[ x=4, y=-9, color=rwth-magenta ]{v1}
			\node[scale = 3.0] at (4, -9.95){ \large \( \bV \)};
			\Vertex[ x=2, y=2, color=rwth-magenta ]{v2}
			\node[scale = 3.0] at (2, 2.9){ \large \( V_1\)};
			\Vertex[ x=8, y=0, color=rwth-magenta ]{v3}
			\node[scale = 3.0] at (6.9, 0.5){ \large \( V_2\)};
			\Vertex[ x=16, y=4, color=rwth-magenta ]{v4}
			\node[scale = 3.0] at (16, 4.9){ \large \( V_3\)};
			\Vertex[ x=18, y=-2, color=rwth-magenta ]{v5}
			\node[scale = 3.0] at (18, -1.0){ \large \( V_N\)};
			\Vertex[ x=12, y=0, color=rwth-blue ]{v6}
			\node[scale = 2.0, draw = rwth-blue, fill=white] at (14, 1.2){ \large \textcolor{rwth-blue}{\( \wt V = \mathrm{Exp}_{[\bV]}^{\Gr}(\wt \Delta)\)}};
		\end{tikzpicture}}
	\caption{Riemannian interpolation in normal coordinates: scheme on \(T_{\bD} \Gr \). Base point \( \bV\) and exact computations \( \{ V_i\}_{i=1}^N\) are shown in magenta; the Lagrangian interpolant \( \wt \Delta \in T_{\bD} \Gr \) of the corresponding Grassman logarithms \( \Delta_i \) and the consequent exponential map \( \wt V \) back to \( \Gr \) are shown in blue. }
	\label{fig:interp_scheme}
\end{figure}
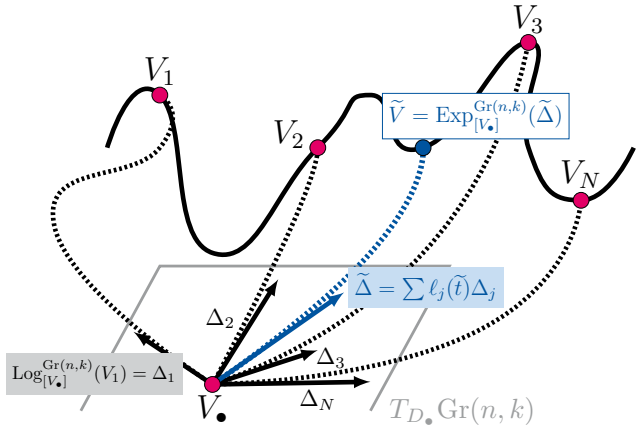

To implement these ideas, we need to introduce the notions of the logarithmic and exponential maps for the Grassmann manifold.
The map between the manifold and the tangent space is given by the logarithmic map \( \mathrm{Log}_{\bD} : \Gr \to T_{\bD} \Gr \) which computes the initial geodesic directions \( \Delta_i \in T_{\bD} \Gr \) starting at the base point \( \bD \) and arriving at the points \( D_i \) in unit time.
Its inverse is the exponential map \( \mathrm{Exp}_{\bD} : T_{\bD} \Gr \to \Gr \) which integrates the geodesic from the base point \( \bD \) with the initial derivative \( \Delta \).
For a given interpolation time \( t \), we can now build the interpolant \( \wt \Delta(t) \) in the tangent space by Lagrange interpolation of the \( \Delta_i \)'s.
This interpolant can now be integrated along the geodesic from \( \bD\) to obtain the desired interpolation \( \wt D(t) \); see Figure~\ref{fig:interp_scheme} for an illustration of the procedure.

% [TODO]{\color{red} if possisble we should make the figure more informative, e.g., by adding the maps Exp and Log, and maybe illustrating the interpolation in the tangent space somehow? In any case, please expand the caption.}

We now proceed with the formal definitions that underpin the above outlined procedure.

\begin{definition}[Tangent space, \cite{bendokat2024grassmann}]
	Let the equivalence class \( [ V ] \in \Gr \) be an element on the Grassmann manifold, corresponding to the projector \( D \). The tangent space at \( [ V ] \), denoted by \( T_{[V]} \Gr \) is given by:
	\begin{equation*}
		T_{ [ V ] } \Gr = \left\{ \Delta \in \ds R^{n \times k} \middle| V^\top \Delta = 0 \text{ for  } V \in [ V ]  \right\}
	\end{equation*}
\end{definition}
Intuitively, the definition states that the tangent space at the subspace \( \mc V \) consists of \( n \times k \) matrices whose image falls into \( \mc V^\perp \), the orthogonal complement of \( \mc V \).
In terms of the corresponding projector \( D \), the tangent space may also be described as:
\begin{equation*}
	T_D \Gr = \left\{ P \in \ds R^{ n \times n } \middle| P = P^\top \text{ with } PD + DP = P \right\}
\end{equation*}
%\mnln{ More common characterization for the tangent vector for the projector is \( P D + D P = P\). Recalling \( D^2 = D\) and multiplying both sides by \(  D \) on the right, one gets \( P D + D P D = P D\), giving \( P D P = 0\). For more involved discussion, see \cite[Prop2.1]{bendokat2024grassmann}.}
where for each \( P \in T_D \Gr \), it holds that \( P V = \Delta \in T_{ [V] } \Gr \) for every \( V \in [ V ] \).

We now proceed to define the logarithmic and exponential map, using Stiefel representatives for the corresponding equivalence classes.
\begin{definition}
	The Grassmann \emph{Logarithm} \( \mathrm{Log}_{ \bV }^{ \Gr } ( V ) : \Gr \mapsto T_{[\bV]} \Gr \) maps the Stiefel representative of \( [ V ] \) to elements \( \Delta \in  T_{[\bV]} \Gr \) such that \( \Delta \) is the direction of the geodesic \( \gamma \) from \( [ \bV ] \) to \( [ V ] \) on \( \Gr \) (that reaches \( [ V ] \) in unit time).
	This map maybe computed using any two respective representatives \( V \in [ V ] \) and \( \bV \in [ \bV ] \) as follows:
	\begin{equation*}
		\begin{array}{l}
			L = V ( \bV^\top V )^{ -1 } - \bV         \\
			U, \xi, W^\top = \texttt{thin\_svd} ( L ) \\
			\Delta = U \arctan \xi W^\top
		\end{array}
	\end{equation*}
	The computational complexity has a bottleneck of the singular value decomposition and is \( \mc O(n k^2)\), \cite{golubMatrixComputations2013}.
	Note that formally we suggest here a so-called Procrustes Logarithm which maintains that the composition of the logarithm and the exponential is the identity map by aligning Stiefel representatives \( \bV \) and \( V \) via obtaining an orthogonal matrix \( O \in \ds R^{ k \times k } \) such that \( O = \underset{ O \in \ds O(k)}{\argmin} \| \bV - V O \|_F \):
	\begin{equation*}
		\begin{array}{l}
			O, \xi', W'^{\top} = \texttt{svd} ( V^\top \bV )                               \\
			\wh V = V O W'^\top                                                            \\
			U, \xi, W^\top = \texttt{thin\_svd} \left( ( I - \bV \bV^{\top}) \wh V \right) \\
			\Delta = U \arcsin ( \xi ) W^\top
		\end{array}
	\end{equation*}
	This computation avoids an explicit inversion of the matrix and maintains the same computational complexity.

\end{definition}

\begin{definition}
	The Grassmann \emph{Exponential} \( \mathrm{Exp}_{ [ \bV ] }^{\Gr} (\Delta) \) is the opposite of the logarithmic map, which takes a base point \( [ \bV ] \) on the manifold and a direction \( \Delta \) in the tangent space \( T_{ [ \bV ] } \Gr  \) and returns a Stiefel representative \( V \)  for the point on the manifold \( [ V ] \) such that \( [ V ] = \gamma(1) \).
	The computation of the exponentialal map may be performed as follows:
	\begin{equation*}
		\begin{aligned}
			 & U, \xi, W^\top = \texttt{thin\_svd}(\Delta) \\
			 & V  = \begin{pmatrix}
				        \bV W & U
			        \end{pmatrix} \begin{pmatrix}
				                      \cos(\xi) \\ \sin(\xi)
			                      \end{pmatrix}W^\top
		\end{aligned}
	\end{equation*}
	The Grassmann exponential has the same computational complexity as the Grassmann logarithm, \( \mc O(n k^2)\).
	Note that one may omit the  factor \( W^\top \) in the formulation of the exponential map, producing a different Stiefel representative of the same equivalence class \( [V]\). Retaining \( W^\top \) allows to maintain the composition of the Procrustes logarithm and the exponent as identity.
\end{definition}

\begin{remark}[Horizontal lift and representatives]
	Conventionally, elements of the tangent space \( T_{[\bV]} \Gr \) are required to exhibit the same dimensionality as elements of the manifold itself.
	However, we here represent elements of \( T_{[\bV]} \Gr \) by the elements in \( \ds R^{n \times k}\) that, by definition, fall on the horizontal space of the Stiefel manifold \( \St \).
	Due to the quotient structure of the Grassmann manifold \( \Gr \), there is a unique lift from the tangent space \( T_{\bD} \Gr \) to the horizontal space (i.e. \( P \mapsto P V \)); see the complete discussion in \cite{bendokat2024grassmann}.

	From a computational standpoint, only Stiefel representatives \( V \in \ds R^{n\times k}\) are stored and processed, making the consideration of the horizontal lift of the tangent space more natural, despite a minor abuse of notation.
	Finally, note that the exponential maps \( \mathrm{Exp}_{[\bV]}^{\Gr} \) should formally return not a Stiefel representative \( V \), but an equivalence class \( [V ]\).
	However, to simplify the discussion below, we consider the actual matrix \( V \) as an output of the exponential map, from which the corresponding \( [V]\in \Gr\) is immediately inferred.
\end{remark}

Using the definitions above, we can compactly write the interpolation scheme as displayed in Algorithm~\ref{alg:interpolationscheme}.

\begin{algorithm}[!t]
	\begin{algorithmic}[1]
		\caption{Riemannian interpolation\label{alg:interpolationscheme}}
		\REQUIRE number of time points \( \{ t_i\}_{i=1}^N \), exact projector estimates \( \{ D_i\}_{i=1}^N \in \mathrm{Gr}_k\) with corresponding Stiefel representatives \( \{ V_i\}_{i=1}^N \in \St \), interpolation time \( t \)
		\STATE choose a base point \( [ \bV] \in \Gr \) with the representative \( \bV \in \St \)
		\STATE \( \Delta_i \gets \mathrm{Log}_{[\bV]}^{\Gr} (V_i) \) for all \( i \in [N] \) {\color{rwth-black-50}\COMMENT{get every direction, \\ \hfill \( \Delta_i \in T_{ [\bV] } \Gr  \)}}
		\STATE \( \wt \Delta(t) \gets \sum_{j=1}^N \ell_j(t) \Delta_j \) \hfill {\color{rwth-black-50}\COMMENT{interpolation; \( \ell_j(t) \)  \\ \hfill is the Lagrange base polynomial}}
		\STATE \( \widetilde V(t) \gets \mathrm{Exp}_{ [\bV]}^{ \Gr }(\wt \Delta(t))\) \hfill{\color{rwth-black-50}\COMMENT{get the interpolated point}}
		\RETURN \( \widetilde V(t) \), \( [ \wt V(t) ] \), \( D(t) = \widetilde V(t) \widetilde V(t)^\top \)
	\end{algorithmic}
\end{algorithm}

\subsection{ Filter interpolation. }

Equipped with the Riemannian interpolation scheme, in this section we now consider the interpolation of graph filters.
Recall that we consider the parametric filter \( H(t) \) built on the first \( k \) eigenvectors with the coefficient vector \( \b h(t) \).
For simplicity, we assume here that the true, correct filter is known and is preserved for all points in times, \( \b h(t) \equiv \b h \).
Alternatively, one can assume a pre-existing or separate oracle that returns \( \b h(t) \) independently of the Grassmanian interpolation.
Our filter is thus given by:
\begin{equation*}
	H(t) = V(t) \diag(\Psi(t) \b h) V^\top(t)
\end{equation*}
where the time dependency in the Vandermond matrix \( \Psi(t) \) is facilitated through the dependence of the spectrum, \( \Psi(t) = \Psi( \b \lambda(t))\).
Our procedure for filter interpolation may now be formulated as follows:
\begin{enumerate}[label={\color{rwth-blue}\arabic*.}]%, itemsep = -5pt, topsep = 0pt]
	\item Assume we are given a set of \( \{t_i, V_i \}\) of eigenspaces in terms of their Stiefel representatives at time points \( t_1, t_2, \dots t_N \) along the graph trajectory \( \mc G(t) \).
	      We can then compute an interpolant \( \wt V(t )\) using the Grassmannian interpolation scheme Algorithm~\ref{alg:interpolationscheme} for an arbitrary time \( t \).
	\item We may also obtain an approximation of the spectrum of the Laplacian \( \Lambda(t) \) through the corresponding eigenvectors \( \wt V(t)\):
	      \begin{equation*}
		      \b{\wt \lambda(t)} \gets  \texttt{eigs}( \wt V(t)^\top L(t) \wt V(t) )
	      \end{equation*}
	      which is the eigenproblem for a \( k \times k \) matrix;
	\item {Note that by Grassmannian interpolation, we obtain only a \emph{representative} \( \wt V(t)\) which may differs from the target basis of the subspace \( \mc V(t)\) by an orthogonal transformation.
	      As a result, the interpolant \( \wt V(t)\) can be misaligned and we  cannot construct the filter \( H(t)\) directly from it.
	      However, the orthogonal transformation \( O \) aligning the representative \( \wt V(t) \) with \( V(t) \) is, in fact, computed during the previous step of spectral approximation:
	      % To remedy this, we note that the eigenvalue approximation \( \b{\wt \lambda(t)} \) above explicitly computes the aligning orthogonal transformation:
	      % {\color{rwth-blue} I don't understand this sentence. What orthogonal transformation; how already obtained?}
	      \begin{equation*}
		      \wt V(t)^\top L(t) \wt V(t)  = O \diag \b{\wt \lambda(t)} O^\top
	      \end{equation*}
	      Then the interpolant filter is defined as follows:
	      \begin{equation*}
		      \wt H(t) = \wt V(t) O \diag( \Psi( \b{\wt \lambda(t)} ) \b h) O^\top \wt V^\top(t).
	      \end{equation*}}
\end{enumerate}

The overall scheme for filter interpolation is summarized in Algorithm~\ref{alg:interpolationschemefilter}.

\begin{algorithm}[htbp]
	\begin{algorithmic}[1]
		\caption{Riemannian interpolation of a graph filter\label{alg:interpolationschemefilter}}
		\REQUIRE number of time points \( \{ t_i\}_1^N \), exact projector estimates \( \{ D_i\}_1^N \in \Gr \) with corresponding Stiefel representatives \( \{ V_i\}_1^N \in \St \), interpolation time \( t \), filter coefficients~\( \b h \)
		\STATE choose a base point \( [\bV] \in \Gr \) with the representative \( \bV \in \St \)
		\STATE \( \Delta_i \gets \mathrm{Log}_{ [\bV] }^{\Gr }( V_i )\) for all \( i \in [N] \) {\color{rwth-black-50}\COMMENT{get every direction, \\ \hfill \( \Delta_i \in T_{ [\bV] } \Gr \)}}
		\STATE \( \wt \Delta(t) \gets \sum_{j=1}^N l_j(t) \Delta_j \) \hfill {\color{rwth-black-50}\COMMENT{interpolation; \( l_j(t) \) \\ \hfill  is the Lagrange base polynomial}}
		\STATE \( \wt V(t) \gets \mathrm{Exp}_{ [\bV] }^{\Gr}( \wt \Delta(t) )\) \hfill {\color{rwth-black-50}\COMMENT{get the interpolated point}}
		\STATE  \( \b{\wt \lambda}, O \gets \texttt{eigs}( \wt V^\top(t) L(t) \wt V(t) ) \) \hfill {\color{rwth-black-50}\COMMENT{get the spectrum}}
		\STATE  \( \wt \Psi(t) \gets \texttt{vandermond}( \b{\wt \lambda}, M) \) \hfill {\color{rwth-black-50}\COMMENT{construct interpolated \\ \hfill Vandermond matrix}}
		\STATE  \( \wt H(t) = \wt V(t) O \diag( \wt \Psi(t) \b h) O^\top \wt V(t)^\top \) \hfill% {\color{rwth-black-50}\COMMENT{get \\ \hfill the interpolated filter}}
		\RETURN \( \wt H(t) \)
	\end{algorithmic}
\end{algorithm}

\subsection{ Error Analysis}
\label{subsec:error}

We now provide an error analysis for the proposed interpolation scheme.
For the task of subspace interpolation, there are multiple possible notions of the interpolation error.

One natural choice is to consider the Grassmann distance between the exact (unobserved) subspace \( [ V(t) ] \) and the interpolant \( [ \wt V ] \):
\begin{equation*}
	\begin{aligned}
		d_{\mathrm{Gr}} ( V(t), \wt V(t) ) & := \| \arccos \Sigma \|_F,                                                  \\
		                                   & \text{  with } \Sigma = \texttt{svdvals} \left( \wt V(t)^\top V(t) \right),
	\end{aligned}
\end{equation*}
which is the length of the geodesic between two points on \( \Gr \) and coincides with the norm of the Riemannian logarithmic map, \(  d_{\mathrm{Gr}} ( V(t), \wt V(t) ) = \| \mathrm{Log}_{V(t)}^{\Gr} (\wt V(t)) \|_F \).

Alternatively, we can consider some matrix-norm of the distance between the projectors \( D(t)\) and \( \wt D(t)\) associated with subspaces \( [V(t)]\) and \( [\wt V(t)]\):
\begin{equation*}
	\begin{aligned}
		d_{F} ( V(t), & \wt V(t))  := d_{F} ( D(t), \wt D(t) )                                        \\
		              & = \| D (t) - \wt D(t) \|_F =  \| V(t) V(t)^\top - \wt V(t) \wt V(t)^\top \|_F
	\end{aligned}
\end{equation*}

By design of Algorithm~\ref{alg:interpolationscheme}, the subspace interpolation error arises from a composition of the Lagrange interpolation error in the tangent space and the sensitivity of the exponential map to perturbations in the tangent space.
\begin{equation*}
	d_{ F }( V( t ), \wt V(t) ) \sim \mathrm{Exp}_{[\bV]}^{ \Gr } \text{ sensitivity } \circ \begin{array}{c}\text{Lagrange} \\ \text{interpolation error}\end{array}
\end{equation*}
We address both parts of this error separately below, before combining them into the final statement.

\subsubsection{Sensitivity of the Exponential Map}
We first establish the sensitivity of the exponential map to perturbations in the tangent space.
\begin{lemma}[Sensivity of the Exponential Map]\label{lem:sens_exp}
	Let \( \Delta, \wt \Delta \in T_{ [\bV] } \Gr \) for a given base point \( [\bV] \); then
	\begin{equation*}
		\begin{aligned}
			\Big\| \mathrm{Exp}_{\bV}^{\Gr} (\Delta) - & \mathrm{Exp}_{\bV}^{\Gr} (\wt \Delta ) \Big\|_F \\ & \le \left( \frac{8C}{\sigma_k(\Delta)} + 2 \right) \| \Delta - \wt \Delta \|_F
		\end{aligned}
	\end{equation*}
	where \( C > 0 \) is a fixed constant and \( \sigma_k(\Delta) \) is the smallest positive singular value of \( \Delta \).
\end{lemma}
\begin{IEEEproof}
	Let \( U, \xi, W^\top = \texttt{thin\_svd} ( \Delta ) \) and   \( \wt U, \wt \xi, \wt W^\top = \texttt{thin\_svd} ( \wt \Delta ) \) with \( V = \mathrm{Exp}_{[\bV]}^{\Gr} (\Delta)\) and \( \wt V = \mathrm{Exp}_{[\bV]}^{\Gr} (\wt \Delta )\). Then, using the triangle inequality, we get
	\begin{equation*}
		\begin{aligned}
			\| V - \wt V \|_F & \le  \left\| \bV \left(  W \cos \xi \, W^\top - \wt W \cos \wt \xi \, \wt W^\top\right) \right\|_F \\ & \phantom{=} + \left\| U \sin \xi \, W^\top - \wt U \sin \wt \xi \, \wt W^\top \right\|_F \\
		\end{aligned}
	\end{equation*}
	Due to the properties of the matrices with orthonormal columns, the first term can be bounded as follows:
	\begin{equation*}
		\begin{aligned}
			\Big\| & \bV \left(  W \cos \xi \, W^\top - \wt W \cos \wt \xi \, \wt W^\top\right) \Big\|_F \\
			       & =  \left\|   W \cos \xi \, W^\top - \wt W \cos \wt \xi \, \wt W^\top \right\|_F     \\
			%    & \le \left\|   W \cos \xi \, W^\top - \wt W \cos \wt \xi \, \wt W^\top \right\|_F   \\
			       & \le \| W \cos \xi ( W - \wt W)^\top \|_F + \| ( W - \wt W) \cos \xi \wt W^\top \|_F \\
			       & \phantom{=} + \| \wt W  ( \cos \xi - \cos \wt \xi ) \wt W^\top \|_F                 \\
			       & \le 2 \, \| W - \wt W \|_F + \| \cos \xi - \cos \wt \xi \|_F
		\end{aligned}
	\end{equation*}
	% {\color{red} Shouldn't there  be additionally a $k$ in front of $\| W - \wt W \|_F$?}
	Applying the similar bound for the second term, we obtain:
	\begin{equation*}
		\begin{aligned}
			\| V - \wt V \|_F & \le 3 \, \| W - \wt W \|_F + \| U - \wt U \|_F \\ & \phantom{=} + \| \sin \xi - \sin \wt \xi \|_F + \| \cos \xi - \cos \wt \xi \|_F,
		\end{aligned}
	\end{equation*}
	% {\color{red} Shouldn't there  be additionally a $k$ in front of $\| W - \wt W \|_F$?}
	which implies that the stability of the exponential map is determined by the stability of the SVD in the Frobenius norm.
	This stability, however, cannot be analysed in a straightforward manner since the SVD is not unique, in general.
	Specifically, let us fix the non-ascending order of the singular values inside the matrix \( \xi \).
	Then the SVD is unique only up to rotations of the singular vectors corresponding to singular values of equal magnitude (i.e. for a single decomposition \( \{ U, \xi, W \} \), we can describe all possible SVDs as \( \{  U Q, \xi, W Q \}  \) where \( Q^\top Q = I \) and \( Q \xi Q^\top = \xi \)).

	As a result, we cannot guarantee that the SVD factors of two close matrices are necessarily close in the Frobenius norm (specifically, singular vector are not necessarily close).
	However, it turns out that the actual \( \mathrm{Exp}_{[\bV]}^{\Gr} (\Delta) \) does \emph{not} depend on the non-uniqueness of the SVD, since \( Q \xi Q^\top  = \xi \) implies \( Q \cos \xi Q^\top = \cos \xi \) and \( Q \sin \xi Q^\top = \sin \xi \) and, as the exponential map is invariant under the allowed rotations of the singular vectors, it is sufficient to find a single set of the SVD factors \( \{ \wt U, \wt \xi, \wt W^\top \} \) that is close in the Frobenius norm.

	For this reason we employ \cite[Theorem 6.4]{stewart1973error} that, for a given \( U, W \), explicitly constructs matrices of singular vectors \( \wt U = U (I +\delta U), \wt W = W (I+\delta W) \) for a given matrix \( \wt \Delta = \wt U \wt \xi \wt W^\top \), such that
	\begin{equation*}
		\begin{aligned}
			 & \| \delta U \|_F = \| U - \wt U \|_F \le C \frac{1}{\sigma_k( \Delta )} \| \Delta - \wt \Delta \|_F, \\
			 & \| \delta W \|_F =\| W - \wt W \|_F \le C \frac{1}{\sigma_k( \Delta )} \| \Delta - \wt \Delta \|_F
		\end{aligned}
	\end{equation*}
	where \( C > 0 \) is an absolute constant and \( \sigma_k(\Delta)\) denotes the smallest positive singular value of the matrix \( \Delta \). %; formally, this bound is a conjecture from the original theorem, \cite{li2024stewart}.
	This is a similar resulat to bounds on the Grassmann distance between singular subspaces, e.g., \( d_{\mathrm{Gr}} (U, \wt U) \le \frac{C}{\sigma_k(\Delta)} \| \Delta - \wt \Delta \|_F \), \cite{vannieuwenhovenConditionNumberSingular2024}.
	In turn, this implies that the singular subspaces of close matrices are indeed close, and we can find an explicit pair of orthonormal bases which are close in the Frobenius norms.

	Note that
	\begin{equation*}
		\begin{aligned}
			\| \cos \xi - \cos \wt \xi \|_F & = \left\| 2 \sin \frac{ \xi +\wt \xi }{2} \odot \sin \frac{ \xi - \wt \xi }{2} \right\|_F \\ & \le \left\| 2 \sin \frac{ \xi - \wt \xi }{2} \right\|_F \le \| \xi - \wt \xi \|_F
		\end{aligned}
	\end{equation*}
	% {\color{red} [shouldn't there be a $k$ in front of the last norm]}
	where \( \odot \) denotes entry-wise matrix multiplication.
	A similar bound can be derived for the sine term.
	Finally, \( \| \xi - \wt \xi \|_F \le \| \Delta - \wt \Delta \|_F \) due to Mirsky's theorem, \cite{stewartPerturbationTheorySingular1998}.

	Combining all the estimates above, we obtain the final bound:
	\begin{equation*}
		\left\| \mathrm{Exp}_{\bV}^{\Gr} (\Delta) - \mathrm{Exp}_{\bV}^{\Gr} (\wt \Delta ) \right\|_F \le \left( \frac{8C}{\sigma_k(\Delta)} + 2 \right) \| \Delta - \wt \Delta \|_F
	\end{equation*}
\end{IEEEproof}

\begin{remark}
	Note that the senstivity of the exponent provides a stricter result than one would need for the interpolation: it states the closeness of the Stiefel representatives of the two subspaces produced by the exponential map, which is a much stronger statement than that the respective subspaces are close.
\end{remark}

\subsubsection{Lagrange Interpolation Error}
We now analyse the interpolation error of our interpolation scheme.
Note that the interpolation accuracy is heavily dependent on the choice of points of the exact computations \( \{ t_i\}_{i=0}^N \); for simplicity we assume \( t_i \in [-1, 1]\).
We may choose the nodes to be equidistant or to follow a specific pattern, such as the Chebyshev points \( t_i = \cos\left( \frac{2i + 1}{2N+2} \pi \right) \), to achieve optimal accuracy and avoid the Runge phenomenon, provided an arbitrary choice of nodes is indeed possible.

To facilitate our analysis, we recall a classic result.
\begin{lemma}[Lagrange Interpolation, \cite{szabados1990interpolation}]\label{lem:lag_error}
	Let \( f \in C^{(N+1)}([-1, 1]) \) and \( \{ t_j \}_{j=0}^N \) be the Chebyshev nodes.
	Then, for the interpolating polynomial \( p_N (t) = \sum_{j=0}^N f(t_j) \ell_j(t) \), the error is bounded by
	\begin{equation*}
		| f(t) - p_N (t) | \le \frac{ 1 }{ 2^{N} (N+1)! } \max_{t \in [-1,1]} \left| f^{(N+1)} (t) \right|
	\end{equation*}\end{lemma}
In the context of Algorithm~\ref{alg:interpolationscheme}, the function \( f \) to be interpolated is the tangent vector \( t \mapsto \Delta(t) = \mathrm{Log}_{[\bV]}^{\Gr} ( V(t) ) \) where \( t \mapsto V(t) \) is a trajectory of corresponding Stiefel representatives.
The combination of the statements above suggests that we need to obtain a bound on the assymptotical behaviour of \( \max | \Delta^{(N+1)}(t) | \) --- provided that the tangent vector trajectory is guaranteed to be smooth enough.
With such a bound we can then obtain an error bound for the eigenspace interpolation process.

\begin{theorem}[Interpolation Error]\label{thm:int_err}
	Let \( \mc V : [-1, 1] \to \Gr \) be a trajectory on the Grassmann manifold with \( \{ t_i \}_{i=0}^N \) being the Chebyshev nodes for the interpolation with corresponding Stiefel representatives \( V_i = V(t_i) \), and \( \bV \in \St \) being a representative of the chosen base point~\([\bV]\).

	We assume that
	\begin{itemize}%[itemsep=-0.5em]
		\item the subspace trajectory \( \mc V(t) \) corresponds to the projector trajectory \( D(t) \) (such that \( V_i V_i^\top = D(t_i)\)) with real analytic entries;
		\item the Riemannian logarithm map is properly defined at every point of the trajectory \( D(t) \) from the base point \( \bV \), i.e. \( V(t)^\top \bV \) is never rank-deficient for any Stiefel representative \( V(t)\);
		      % \item eigenspace trajectory \( \mc V(t) \) contains a smooth trajectory of Stiefel representatives \( V(t) \in C^{(N+1)}([-1,1]) \) such that all exact computations \( V_i \in V(t) \) and \( V(t)^\top \bV \) does not have neither \( 0 \) singular values or singular vectors corresponding to the same singular values for every \( t \);
		\item the corresponding trajectory of the tangent vectors \( \Delta(t) \) is never rank-deficient, i.e. \( \sigma_k(\Delta(t))  { \ge \sigma_{\min} }> 0 \) uniformly for all \( t \);
		      %\item and the higher-order derivatives of \( V(t) \) have controlled growth, \( \| {V^\top}^{(M)} \|_F \le \alpha \| {V^\top}^{(M-1)} \|_F \), for some constant \( \alpha > 0 \).
	\end{itemize}
	Then, the interpolation error between the exact eigenspace \( V(t) \) and the interpolant \( \wt V(t) \) defined in Algorithm~\ref{alg:interpolationscheme}
	is bounded by
	\begin{equation*}
		\begin{aligned}
			d_F(V(t), \wt V(t)) & = \| V(t) V(t)^\top - \wt V(t) \wt V(t)^\top \|_F \\ & \le  4 \left( \frac{4C}{ { \sigma_{\min} }
				% {\sigma_k(\Delta(t))}
			} +1  \right) \frac{\max_{\tau \in [-1, 1]} \| \Delta^{(N+1)}(\tau) \|_F}{2^N (N+1)!}
		\end{aligned}
		%\lesssim  \frac{1}{\sigma_k} \left( \frac{ \alpha k^2 }{2} \right)^{N-1}
	\end{equation*}
	for any \( N \in \ds N\) .
	% {\color{red} [Why not $\wt V(t)$? The interpolant also depends on time. Same in the proof.]}
\end{theorem}

\begin{IEEEproof}
	To obtain the desired bound, we mainly need to combine the lemmas above.

	First, note that
	\begin{equation*}
		\begin{aligned}
			\| V(t) V(t)^\top - \wt V(t) \wt V(t)^\top \|_F \le \| V(t) ( V(t) - \wt V(t) )^\top \|_F \\ + \| (V(t) - \wt V(t)) \wt V(t)^\top \|_F = 2 \, \| V(t) - \wt V(t) \|_F
		\end{aligned}
	\end{equation*}
	Hence, one can show
	\begin{equation*}
		\begin{aligned}
			d_F(V(t), \wt V(t)) & \le 2 \, \| V(t) - \wt V(t) \|_F
			\\ & = 2 \left\| \mathrm{Exp}_{[\bV]}^{\Gr}(\wt \Delta(t)) - \mathrm{Exp}_{[\bV]}^{\Gr}(\Delta(t)) \right\|_F  \\
			                    & \le 4 \left( \frac{4C}{
				{ \sigma_{\min} }
				%  \sigma_k(\Delta(t))
			} +1  \right) \| \wt \Delta(t) - \Delta(t) \|_F %\le                                      \\
			%			                                                 & \le 2 (2k+1) \left( \frac{2C}{\sigma_k(\Delta(t))} +1  \right) \frac{\max_{\tau \in [-1, 1]} \| \Delta^{(N+1)}(\tau) \|_F}{2^N (N+1)!}
		\end{aligned}
	\end{equation*}
	The bound on the distance between the true value of the tangent vector \( \Delta(t) \) and the interpolant \( \wt D \) may now be derived via Lemma~\ref{lem:lag_error}.

	To apply the lemma we need to guarantee the necessary smoothness of the tangent vector trajectory \( \Delta(t) \), such that the \( (N+1) \)-th derivative exists.
	In both derivations of the logarithm above, the problematic aspect for smoothness is the computations of the singular value decomposition of matrices \( V^\top \bV \) or \( \bV^{\top} V \), respectively.
	Note that by an immediate corollary from \cite[Theorem 1]{rellich1969perturbation}, there exists an analytic eigendecomposition \( D(t) = V(t) V(t)^\top \) where \( V(t)\) is real analytic entry-wise.

	Using our second assumption that \( \bV^\top V(t) \) is never rank deficient, it is hence invertible and analytic so that the matrix \( L = V(t) (\bV^\top V(t))^{{-1}} - \bV \) is analytic as well.
	Using again the results from \cite{rellich1969perturbation}, we can obtain an analytic eigendecomposition for \( L L^\top = U(t) \xi^2(t) U(t)^\top \) where \( U(t)\), \( \xi^2(t)\) (and, hence, \( \xi(t)\)) are real analytic entry-wise.
	This implies analiticity for the singular values and left singular vectors used in the logarithm.

	It is now enough to note that by the last assumption, \( \sigma_k (\Delta(t))>0\) with \( \Delta(t) = U(t) \arctan \xi(t) W(t)^\top\), hence \( \xi(t) \) is invertible and \(\xi^{-1}(t)\) is analytic.
	Then, for \( L = U(t) \xi(t) W(t)^\top \), we can derive \( W(t) = L U(t) \xi^{-1}(t) \) where all factors are well-defined and analytic.

	As a result, \( \Delta(t) = U(t) \arctan \xi(t) W(t)^\top \) is guaranteed to be smooth enough to apply the lemma and the error can be propagated as follows:
	\begin{equation*}
		\begin{aligned}
			d_F(V(t), \wt V(t)) & \le 4 \left( \frac{4C}{
				{ \sigma_{\min} }
				% {\sigma_k(\Delta(t))
			} +1  \right) \| \wt \Delta - \Delta (t) \|
			\\
			                    & \le 4 \left( \frac{4C}{ {\sigma_{\min} }
				% {\sigma_k(\Delta(t))
			}+1  \right) \frac{\max_{\tau \in [-1, 1]} \| \Delta^{(N+1)}(\tau) \|_F}{2^N (N+1)!}
		\end{aligned}
	\end{equation*}
\end{IEEEproof}

\begin{remark}[Relaxation analyticity]
	The assumptions of Theorem~\ref{thm:int_err} provide a higher level of regularity than required for the \( C^{(N+1)}\) class needed for the interpolation error.
	We note that one can relax the analyticity of \( D(t) \) with a trade-off such as one can allow \( D(t) \in C^{(N+1)}\) while \( V^{*\top} D(t) \bV \) does not have coinciding eigenvalues (i.e. one cannot allow for \([\bV] \in [V(t)] \)).
\end{remark}

\begin{remark}[Rank-deficiency assumptions]
	The statement of Theorem~\ref{thm:int_err} asks for full-rank tangent vector \( \Delta (t) \) and full-rank matrix \( V(t)^\top \bV \).
	Both assumptions are necessary: note that \( \Delta(t) \) can generally have a non-trivial kernel up to \( \Delta(t) = \mathrm{Log}_{\bV}^{\Gr} (\bV ) = 0 \) since \( \|  \mathrm{Log}_{\bV}^{\Gr} (\bV ) \| = d_{\mathrm{Gr}}(\bV, \bV) = 0 \).
	% {\color{red} [I see the math, but it still feels non-intuitive.]}
	Possible rank deficiency of the \( V(t)^\top \bV \) matrix is more nuanced.
	Using the Procrustes logarithm, we can avoid inverting the matrix \( V(t)^\top \bV \).
	However, in the rank-deficient case the analyticity of the right singular vectors of \( V(t)^\top \bV \) and \( (I - \bV \bV^\top) \wh V \) becomes non-trivial.
\end{remark}

\begin{remark}[Bound on the norm of the higher-order derivative]
	\label{rem:norm}
	The error bound from Theorem~\ref{thm:int_err} is scaled by the norm of the higher-order derivative of the tangent vector \( \| \Delta^{(N+1)}(\tau) \|_F \) which is not constructive since it calls for higher-order characterization of the trajectory of tangent vectors.
	One can bound this norm explicitly by assuming that the change in the tangent vector \( \Delta (t) \) is concentrated at singular values and such singular values have derivatives of decaying norms.
	In particular, assume that \( \| U^{(M)}(t) \|_F \le \beta \| \xi^{(M)}(t) \|_F \) and \( \| W^{(M)}(t) \|_F \le \beta \| \xi^{(M)}(t) \|_F \) for all \( M \ge 1\) and \( \| \xi^{(M)}(t) \|_F \le \alpha^M \| \xi(t) \|^M_F \) for some \( \alpha, \beta > 0 \) and every derivative of order \(M\), \( 1 \le M \le N+1 \), uniform in time.
	This assumption implies that:
	\begin{equation*}
		\small
		\begin{aligned}
			\| \Delta^{(M)} \| & = \left\| \sum_{\substack{k_1, k_2, k_3 \ge 0   \\ k_1 + k_2 + k_3 = M}} \binom{M}{k_i} U^{(k_1)} \xi^{(k_2)} W^{(k_3)\top}\right\| \\ & \le  \sum_{\substack{k_1, k_2, k_3 \ge 0 \\ k_1 + k_2 + k_3 = M}} \binom{M}{k_i} \| U^{(k_1)} \| \cdot \|  \xi^{(k_2)} \| \cdot \|  W^{(k_3)\top}\|  \\
			                   & \le \beta^2 \sum_{\substack{k_1, k_2, k_3 \ge 0 \\ k_1 + k_2 + k_3 = M}} \binom{M}{k_i} \| \xi^{(k_1)} \| \cdot \|  \xi^{(k_2)} \| \cdot \|  \xi^{(k_3)}\| \\ & \le \beta^2 \alpha^M \| \xi \|^M \sum_{\substack{k_1, k_2, k_3 \ge 0       \\ k_1 + k_2 + k_3 = M}} \binom{M}{k_i} = \beta^2 (3\alpha)^M \| \xi \|^M
		\end{aligned}
	\end{equation*}
	Note that \( \| \xi \| = \| \Delta \| = \| \mathrm{Log}_{[\bV]}^\Gr V(t) \| = d_{\mathrm{Gr}} (V(t), \bV)\) by the definition of the Riemannian logarithmic map.
	As a result,
	\begin{equation*}
		\begin{aligned}
			d_F(V(t), \wt V) \le \wt C \frac{ (3\alpha)^{N+1} \left[ \max_\tau d_{\mathrm{Gr}} (V( \tau ), \bV)\right]^{N+1} } { 2^N (N+1)! } \\  \lesssim \left( \frac{3e \alpha \cdot \max_\tau d_{\mathrm{Gr}} (V( \tau ), \bV)}{2(N+1)} \right)^N
		\end{aligned}
	\end{equation*}
	In the assumptions of Theorem~\ref{thm:int_err}, \( d_{\mathrm{Gr}} ( V(\tau), \bV ) \) is well-defined and analytic, hence \( \max_{ \tau \in [-1, 1] } d_{\mathrm{Gr}} (V( \tau ), \bV) < \infty \).
	As a result, the inequality above states fast convergence in the order of interpolation for the error independently of the magnitudes of \( \beta \) and \( \alpha \) suggesting that such an error bound holds for all functions with the at-most exponential growth of the norm of higher-order derivatives.

\end{remark}

\section{Illustrative application I: Similarity correction and CSBM }
\label{sec:csbm}

% Parametric graph families are reasonably common (f.i. electrical grids, transportation, citation or sensor networks) in natural systems; at the same time, one can also suggest a number of scenarios where the parametric family is induced.
% In this section we suggest a framework where the evolving topology is tuned to reflect the changing homophily of the nodes.
% We then provide a theoretical result connecting the similarity correction and the spectral information of the parametric graph shift operator \( S(t)\); we use the suggested approach to illustrate the performance of the interpolation approaches of Algorithms~\ref{alg:interpolationscheme} and \ref{alg:interpolationschemefilter} using this similarity correction.

Parametric families of graphs—collections of graphs indexed by an exogenous parameter \(t\)—arise naturally in domains where connectivity patterns evolve with time, operating conditions, or scale.
Examples include electrical grids responding to demand and contingency, transportation networks subject to seasonal schedules and disruptions, citation graphs growing as literature expands, and sensor networks adapting to environmental stimuli.
Beyond such naturally evolving systems, parametric dependence can also be induced by modeling or algorithmic choices: varying sparsification thresholds, diffusion scales in kernel constructions, or regularization weights in graph learning each produce a structured family \(\{S(t)\}_{t\in\mathcal{T}}\) of graph shift operators that are comparable across \(t\).

In this section, we introduce a framework that couples the evolution of topology to changes in vertex homophily.
By homophily we mean the tendency of vertices with similar attributes to be more strongly connected.
When vertex attributes or their relevance vary with \(t\), the induced similarity relations should be reflected in the graph structure.
We formalize this intuition by designing a similarity correction that updates edge weights so that the parametric graph shift operator \(S(t)\) remains consistent with the instantaneous similarity landscape while preserving comparability across the parameter domain.
Throughout, \(S(t)\) can be instantiated as an adjacency matrix, (normalized) Laplacian, or another standard graph shift operator, and we assume a fixed vertex set with edge weights modulated by \(t\).

We use this similarity-corrected family of graphs to illustrate and assess the performance of the interpolation approaches implemented in Algorithms~\ref{alg:interpolationscheme} and \ref{alg:interpolationschemefilter}.
We demonstrate that the correction improves spectral alignment across \(t\), which in turn reduces interpolation error for both operator- and filter-centric schemes.
The examples in this section illustrate these effects and highlight the practical value of coupling topology evolution with homophily-aware similarity correction.

\subsection{Homophily}
Let \( S \) be a fixed shift operator describing a static network of interactions in which each vertex has a time dependent feature vector \( \b x_i(t) \in \ds R^d\).
Thus, rather than explicitly modeling a time-varying topology, we study how a time-varying feature space interacts with a fixed base connectivity.
In graph signal processing (GSP) and graph neural networks (GNNs), homophily—the tendency of connected vertices to have similar attributes—and its counterpart heterophily are commonly used to characterize the regime in which filters or learned models perform well: low-pass filters (and many GNN architectures) are typically advantageous on homophilic graphs where signals are smooth over edges, whereas high-pass filters and architectures tailored to heterophily benefit when connected vertices are dissimilar.
When vertex attributes \( \{ \b x_1(t), \ldots \b x_n(t) \} \) evolve over time, the degree of homophily induced by the features along the fixed edges may drift, which can degrade performance if the filter or model is mismatched to the current regime.

To mitigate such effects, we propose a similarity correction that adapts edge weights to the instantaneous feature similarity while preserving the base topology. Let \(w([v_i,v_j])\) be the original weight on edge \([v_i,v_j]\). We now define a similarity corrected weight:
\begin{equation*}
	\begin{aligned}
		w^{new} ([v_i v_j]) = w([v_i, v_j]) \cdot \cos (\b x_i(t),  \b x_j(t)) \\ = w([v_i v_j]) \cdot \frac{\b x_i(t)^\top \b x_j(t)}{\|\b x_i(t)\| \cdot \|\b x_j(t)\|}
	\end{aligned}
\end{equation*}
This construction induces a parametric family \(S(t)\) whose spectrum and associated frequency notions track the evolving homophily of the features. Under mild regularity assumptions on \(t\mapsto \b x_i(t)\), classical perturbation results imply that the spectral structure of \(S(t)\) varies smoothly, which is advantageous for interpolation and transport of filters across time.

Note that while it is common for features to be non-negative, \( \b x_i(t) > 0 \), the above construction can also be adjusted to use the absolute value of the similarity to avoid negative weights.
Similarly, the usage of a different similarity measure instead of cosine similarity is also possible depending on the specific application and the nature of the data.
Below, we illustrate the effects of this similarity correction on a random graph model endowed with time-varying vertex features.

\subsection{Contextual Stochastic Block Model}

The Contextual Stochastic Block Model (CSBM) augments the classical Stochastic Block Model (SBM) by associating feature vectors to vertices and linking edge formation to both community structure and attribute similarity \cite{deshpande2018contextual}.
To keep notation simple, we consider a two-block instance with a fixed vertex set partitioned into clusters of sizes \(n_1\) and \(n_2\) (with \(n_1+n_2=n\)).
Edges are generated independently: an intra-cluster edge appears with probability \(p>0\), and an inter-cluster edge with probability \(q>0\).
Each vertex \(v_i\) in community \(c\in\{1,2\}\) is endowed with a time-varying feature vector \(\b x_i(t)\in\mathbb{R}^d\) sampled independently from a Gaussian distribution \(\mathcal{N}\!\big(\b\mu_c(t),I_d\big)\), where the mean \(\b\mu_c(t)\) encodes temporal context and \(I_d\) is the identity covariance.

We apply the similarity correction introduced above by modulating edge weights with the cosine similarity of contemporaneous features.
For an unweighted base CSBM, this transforms the adjacency at time \(t\) into a weighted matrix \(A(t)\) whose entries are the product of the Bernoulli edge indicators and the cosine similarity \(\cos\big(\b x_i(t),\b x_j(t)\big)\).
The following lemma characterizes the expected adjacency and shows that the block structure of the SBM is preserved while being adaptively scaled by feature similarity.

\begin{lemma}[Expected temporal CSBM]\label{lem:csbm}
	For sufficiently large \(d\), the expected adjacency matrix of the similarity-corrected CSBM admits the approximation
	\[
		\mathbb{E}\,A(t)\approx
		\begin{bmatrix}
			p\,\kappa_{11}(t)\,\big[\b 1_{n_1}\b 1_{n_1}^\top\!-\! I_{n_1}\big] & q\,\kappa_{12}(t)\,\b 1_{n_1}\b 1_{n_2}^\top                        \\
			q\,\kappa_{12}(t)\,\b 1_{n_2}\b 1_{n_1}^\top                        & p\,\kappa_{22}(t)\,\big[\b 1_{n_2}\b 1_{n_2}^\top\!-\! I_{n_2}\big]
		\end{bmatrix},
	\]
	where \(\b 1_k\) denotes the \(k\)-dimensional all-ones vector, and for \(a,b\in\{1,2\}\) we define
		{\footnotesize
			\begin{equation*}
				\kappa_{ab}(t)=\cos_d\!\big(\b\mu_a(t),\b\mu_b(t)\big):=\frac{\b\mu_a(t)^\top\b\mu_b(t)}{\sqrt{\|\b\mu_a(t)\|^2+d}\,\sqrt{\|\b\mu_b(t)\|^2+d}}
			\end{equation*}}
\end{lemma}

\begin{IEEEproof}
	Consider two vertices \(v_i\) and \(v_j\) in the first cluster. The expected adjacency entry after similarity correction is
	\[
		\mathbb{E}\,A_{ij}(t)\;=\;p\,\mathbb{E}_{\b x_i,\b x_j\sim\mathcal{N}(\b\mu_1(t),I_d)}\!\left[\cos\big(\b x_i(t),\b x_j(t)\big)\right].
	\]
	Write the cosine similarity as
	\[
		\cos\big(\b x_i,\b x_j\big)\;=\;\frac{\b x_i^\top\b x_j}{\|\b x_i\|\,\|\b x_j\|}.
	\]
	By independence, \(\mathbb{E}[\b x_i^\top\b x_j]=\b\mu_1(t)^\top\b\mu_1(t)=\|\b\mu_1(t)\|^2\). For Gaussian \(\b x\sim\mathcal{N}(\b\mu,I_d)\), \(\|\b x\|^2\) follows a noncentral \(\chi^2\) distribution with mean \(\|\b\mu\|^2+d\) and concentrates around its mean as \(d\) grows. Using standard concentration (e.g., via Lindeberg-type arguments for sums of independent components) and a smoothness approximation for reciprocal norms, we obtain
	\[
		\mathbb{E}\left[\frac{1}{\|\b x_i\|\,\|\b x_j\|}\right]\;\approx\;\frac{1}{\sqrt{\|\b\mu_1(t)\|^2+d}\,\sqrt{\|\b\mu_1(t)\|^2+d}}.
	\]
	Combining the numerator and denominator approximations yields
	\begin{eqnarray*}
		\mathbb{E}\,\cos(\b x_i(t),\b x_j(t)) &\approx & \dfrac{\|\b\mu_1(t)\|^2}{\sqrt{\|\b\mu_1(t)\|^2+d}\,\sqrt{\|\b\mu_1(t)\|^2+d}}\\
		&=&\cos_d\big(\b\mu_1(t),\b\mu_1(t)\big).
	\end{eqnarray*}
	The same reasoning applies to pairs within the second cluster, giving \(\cos_d\!\big(\b\mu_2(t),\b\mu_2(t)\big)\), and to cross-cluster pairs, for which \(\mathbb{E}[\b x_i^\top\b x_j]=\b\mu_1(t)^\top\b\mu_2(t)\), yielding \(\cos_d\!\big(\b\mu_1(t),\b\mu_2(t)\big)\). Since self-loops are excluded, diagonal entries are zero, which explains the \(-I_{n_c}\) terms. Stacking blocks produces the stated approximation.
\end{IEEEproof}

\begin{remark}
	Lemma~\ref{lem:csbm} formalizes how similarity correction preserves and accentuates the community structure in expectation: intra-block weights scale with \(\kappa_{11}(t)\) and \(\kappa_{22}(t)\), while inter-block weights scale with \(\kappa_{12}(t)\). As \(\b\mu_1(t)\) and \(\b\mu_2(t)\) drift, the effective homophily of the graph changes—inter-block connections become discounted when \(\kappa_{12}(t)\) decreases, and intra-block coherence is reinforced when \(\kappa_{11}(t)\), \(\kappa_{22}(t)\) increase. This matters for low-pass filtering and spectral methods: the expected adjacency (and likewise Laplacian) remains a rank-two perturbation of a block-constant matrix whose leading eigenvectors span the subspace generated by the community indicators. Consequently, spectral clustering and graph filters that rely on smoothness continue to reflect the latent communities even as features evolve over time \cite{JMLR:v22:20-391}. If signed similarities arise (e.g., negative cosine values), one may either clamp or re-map similarities to nonnegative weights, or employ signed Laplacians to explicitly model heterophily, depending on the intended downstream task.
\end{remark}

\subsection{ Numerical example }

After establishing the theoretical motivation of the similarity correction above, we use the CSBM model to demonstrate the performance of the interpolation routines, Algorithms~\ref{alg:interpolationscheme} and \ref{alg:interpolationschemefilter}.
We consider a two cluster CSBM model, \( n_1 = 200 \) and \( n_2 = 400 \), with close inter- and intra-cluster connection probabilities, \( p = 0.45\) and \( q= 0.4\), which suggest a much less pronounced cluster structure without the similarity correction.
The vertex features are sampled from \( \mc N( \mu_1(t) \b 1,  \sigma I_d )\) and \( \mc N( \mu_2(t) \Pi (t) \b 1, \sigma I_d )\) respectively, where \( \Pi(t) \) is a time-dependent rotation matrix controlling the expected cosine similarity between features, Lemma~\ref{lem:csbm}.

\begin{figure*}[!t]
	\begin{center}
		\includegraphics[width=0.95\textwidth]{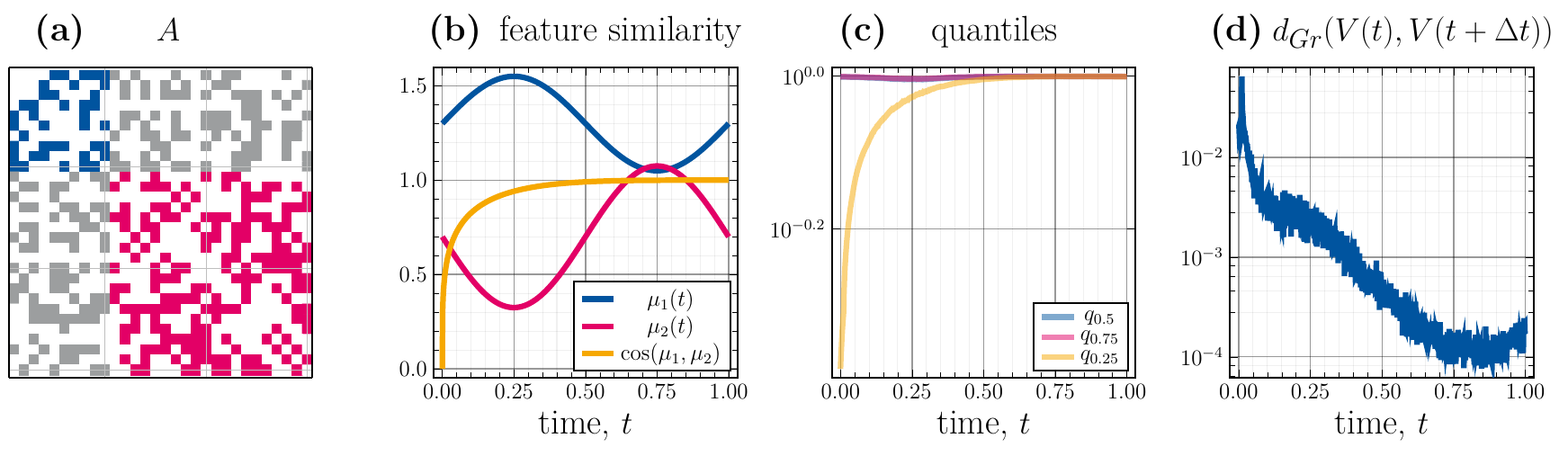}
	\end{center}
	\vspace{-20pt}
	\caption{ CSBM setup parameters: (a) adjacency matrix with colors corresponding to different cluster and inter-cluster edges; (b) evolution of the features means \( \mu_1 \) and \( \mu_2 \) with the cosine similarity of the corresponding vectors; (c) quantiles of the edge weights with similarity correction illustrating changing topology of \(S(t)\); (d) distance between neighbouring low-frequency subspaces along the trajectory \( \mc V(t) \), \( k = 5 \) supporting the fact that \( \mc V(t) \) is slowly changing trajectory suitable for interpolation. }\label{fig:csbm_plot}
\end{figure*}

We illustrate all the settings of the model in Figure~\ref{fig:csbm_plot} on a small sample network (\( n_1 = 10 \), \( n_2 = 20 \)) starting from the block structure (Figure~\ref{fig:csbm_plot}a).
The evolution of the features means \( \b \mu_1(t) \) and \( \b \mu_2(t) \) is set up in a way that the rotation \( \Pi(t) \) aligns two vectors gradually upon approaching \( t = 1\), Figure~\ref{fig:csbm_plot}b.
Figure~\ref{fig:csbm_plot}c demonstrates the evolution of the distribution of edge weights in \( \mc G(t) \) modified by the similarity correction.
Most notably, the trajectory of the first quartile implies that there is a sufficient change between the overall topology of the graph along the time.
The rightmost pane characterizes the evolution of the extremal subspace \( \mc V(t) \) suggesting that the target subspace both is not conserved and changes moderately enough to attempt the interpolation.

\begin{figure*}[!b]
	\begin{center}
		\includegraphics[width=0.95\textwidth]{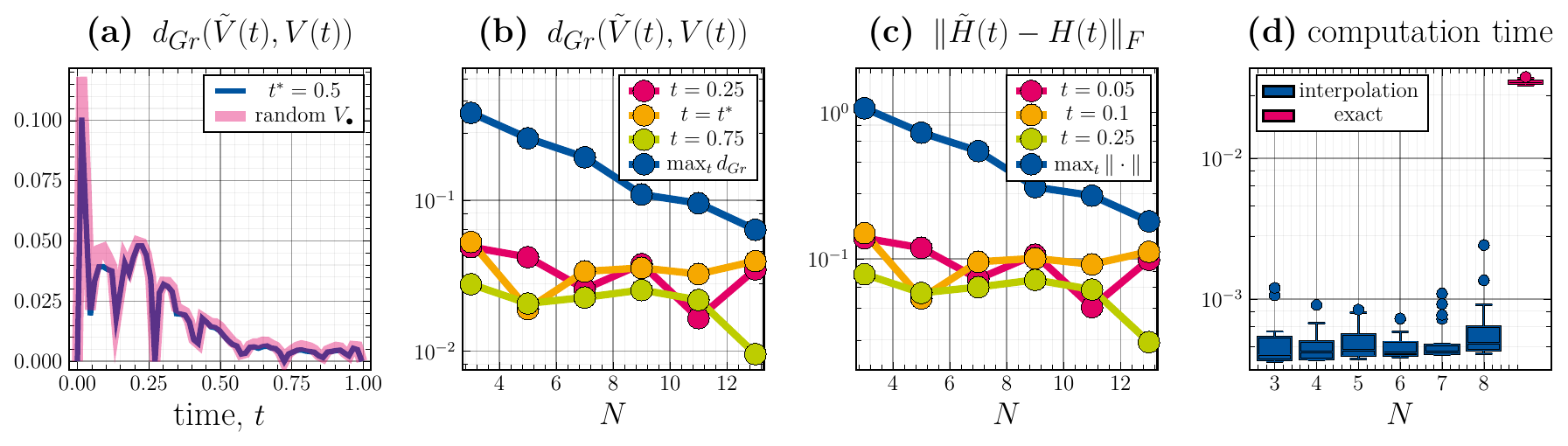}
	\end{center}
	\vspace{-20pt}
	\caption{Interpolation results for CSBM: (a) distance between the exact computation of the extremal subspace \( V(t) \) and the interpolant \( \wt V(t) \) for base point \( \bV \) chosen on the trajectory \( \mc V(t) \) and randomly (dips relate to Chebyshev nodes, \( N = 10 \)); (b) decreasing subspace interpolation error, maximal and at different points times (error changes depending on the location of interpolation nodes), vs number of interpolation nodes \( N \); (c) decreasing filter interpolation error, maximal and at different points times, vs number of interpolation nodes \( N \); (d) gains in the computation time between the filter interpolation and the exact computation. }\label{fig:csbm_res}
\end{figure*}

We then provide the results of the eigenspace interpolation for \( k = 8 \) (Algorithm~\ref{alg:interpolationscheme}) and the filter interpolation (Algorithm~\ref{alg:interpolationschemefilter}) for a sample constant tuple of filter coefficients \( \{ h_i = 2^{-i} \}_{i=0}^4 \) on Figure~\ref{fig:csbm_res}.
% {\color{red} vats? I don't understand the previous sentence.}
Figure~\ref{fig:csbm_res}a indicates that the error of the subspace interpolation is not uniform in time.
Moreover, the error is not largely affected by the choice of the base point \( \bV \), and is characterized by (a) the trajectrory's rate of change (see, for instance, Figure~\ref{fig:csbm_plot}d) and (b) by the closeness to the exact computation points (in this simulation, we opt for the Chebyshev nodes).

Specifically, our simulations suggests that slower changing segments of the trajectory \( \mc G(t)\) are generally better interpolated.
Similarly, the moments of time \( t \) closer to the interpolation nodes \( \{ t_i \}_{i=1}^N \) exhibit comparatively smaller interpolation error.
Consequently, if we increases the number of the interpolation nodes \( N \), then the interpolation error at a fixed time \( t \) is not guaranteed to strictly decrease since its relative position to the interpolation nodes changes (see green, red, and orange lines on Figure~\ref{fig:csbm_res}b).
Nevertheless, the maximal (over time \( t \)) interpolation error on the segment successfully decreases with the number of interpolation nodes \( N \) for both subspaces and filter (Figure~\ref{fig:csbm_res}b,c).
Finally, the interpolation framework expectedly requires sufficiently smaller computation time even for a moderate scale system with \( n = 600\), Figure~\ref{fig:csbm_res}d.
% {\color{red} we need some kind of baseline to compare the against what happens if we do not do interpolation? Also some kind of filtering task that is evaluated with interpolation w/o  interpolation and with constant eigenvectors? Fonts in figures are too big and should not be times roman I think. Give panels proper labels (a), (b), (c), (d) so we can refer to them in the text easier, left right is not so readable/direct.}

\section{Illustrative Application II: Dot Product Graphs and Topology Inference for Classification Tasks }
\label{sec:dpg}

In this section we investigate a scenario where we induce a family of graphs for a given static network.
Specifically, we study a setting in which a parametric family graphs is constructed from a given static network to improve vertex classification.
The central idea is to leverage low-rank latent embeddings derived from the original topology to define dot product graphs (DPGs) whose connectivity can be tuned by a threshold parameter.
This induces a filtration of graphs that trades density against homophily and can yield a topology more conducive to label smoothing and propagation.
Crucially, exploring this family efficiently—and identifying near-optimal thresholds—hinges on the filter interpolation methods developed in this work.

\subsection{Dot product graphs (DPGs)}

Let \( \mathcal{V}=\{v_1,\ldots,v_n\} \) and \( f:\mathcal{V}\to\mathbb{R}^d \) be a vertex embedding. Given a threshold \( \delta\in\mathbb{R} \), the dot product graph places an edge between \(v_i\) and \(v_j\) if and only if \( f(v_i)^\top f(v_j)>\delta \). To encode edge strength while maintaining a thresholded support, we consider the weight
\[
	w([v_i,v_j])\;=\;f(v_i)^\top f(v_j)-\delta,
\]
and, if nonnegativity is desired, the soft-thresholded form \( w([v_i,v_j])=\max\{f(v_i)^\top f(v_j)-\delta,0\} \) (Heaviside-based soft-thresholding). This choice preserves interpretability: edges activate when the latent similarity surpasses \(\delta\), and the weight reflects the margin.

To infer DPG structure from a pre-existing network, we adopt a two-sided spectral embedding that captures the dominant latent geometry of the adjacency:
\begin{enumerate}[label={\color{rwth-blue}\arabic*.}]
	\item Fix \(d\in\mathbb{N}\). For the given graph \(\mathcal{G}\), let \(A\in\mathbb{R}^{n\times n}\) denote its adjacency matrix.
	\item Compute a singular value decomposition \(A=USV^\top\). Let \(U_d\), \(S_d\), and \(V_d\) be the leading \(d\)-dimensional factors, such that \(A_d=U_d S_d V_d^\top\) is the Frobenius-optimal rank-\(d\) approximation.
	\item Define left and right embeddings \( f(v_i)=\big[U_d\sqrt{S_d}\big]_{i,\bullet} \) and \( g(v_i)=\big[V_d\sqrt{S_d}\big]_{i,\bullet} \).
	\item Construct a symmetric DPG by declaring \([v_i,v_j]\in\mathcal{E}\) if and only if
	      \[
		      \min\big\{f(v_i)^\top g(v_j),\,f(v_j)^\top g(v_i)\big\}\;>\;\delta,
	      \]
	      with edge-weight
	      \[ w([v_i,v_j])=\min\big\{f(v_i)^\top g(v_j),\,f(v_j)^\top g(v_i)\big\}-\delta. \]
\end{enumerate}
This two-sided construction is natural because \(A_d=L R^\top\) with \(L=U_d\sqrt{S_d}\) and \(R=V_d\sqrt{S_d}\), so the bilinear form \(f(v_i)^\top g(v_j)\) approximates \(A_{ij}\) in the latent space.

\begin{remark}
	Note that for a symmetic \( A \), one can always find a symmetric best rank-\(d\) approximation \( A_d = L R^\top \) guaranteeing \(f(v_i)^\top g(v_j) = f(v_j)^\top g(v_i)\).
	As a result, the condition of the edge \( [v_i, v_j] \) appearing in the corresponding symmetric DPG is simplified to \(f(v_i)^\top g(v_j)>\delta\).
	% Even when \(A\) is symmetric, \(A_d=L R^\top\) is not necessarily symmetric, hence \(f(v_i)^\top g(v_j)\neq f(v_j)^\top g(v_i)\) in general. Thresholding either quantity alone can produce a directed DPG. To avoid losing symmetry (and diagonalizability in downstream operators), we suggest symmetrization via wrappers such as \(\min(\cdot,\cdot)\), \(\max(\cdot,\cdot)\), or the average \(\tfrac{1}{2}\big(f(v_i)^\top g(v_j)+f(v_j)^\top g(v_i)\big)\) \cite{marencoOnlineChangePoint2022}. For symmetric graphs that are well-approximated by a positive semidefinite model, an eigendecomposition \(A=V\Lambda V^\top\) with \(f(v_i)=\big[V_d\sqrt{\Lambda_d^+}\big]_{i,\bullet}\) offers an alternative, but care is needed if negative eigenvalues are present.
\end{remark}
% {\color{red} I would disagree with this remark, the best low-rank approximation can always be chosen to be symmetric. If the original matrix is symmetric, then the SVD and EVD coincide up to sign flips (you have to move the sign out of the eigenvalues and into one of the singular vectors). If the best-rank approximation is not unique, then you can always find a symmetric one among the best approximations.}

\begin{proposition}
	The procedure above induces a parametric family \(\mathcal{G}(\delta)\) from a static graph \(\mathcal{G}\) in which \(\delta\) defines a filtration of edges by latent similarity. The resulting adjacency of the DPG resembles a thresholded, low-rank approximation of \(A\), effecting a principled re-wiring toward the dominant latent structure.
\end{proposition}

\subsection{Classification via low-pass filter}

We consider a vertex classification task with partial labels.
In the binary setting, each vertex carries a scalar signal \(x_i\in\{-1,1\}\).
We assume that we observe a subset of these labels and use them to assemble the vector of known entries \(\b x^{kn}\), where we set missing entries to \(0\).

Following \cite{sandryhailaDiscreteSignalProcessing2013}, we learn a spectral low-pass filter that propagates labels smoothly over the graph.
Let \(V\) denote the eigenvector matrix of the chosen shift operator, and let \(\Psi\) the Vandermonde matrix of the filter basis (e.g., \(\Psi_{ij}=\lambda_i^{j-1}\) for polynomial filters).
We seek \(\b h\) so that the filtered training signal has the same sign as \(\b x^{kn}\) on labeled vertices.
The learning problem can now be formulated as
\[
	\min_{\b h}\;\Big\| \diag(\b x^{kn})\,V\,\diag(\Psi\b h)\,V^\top\,\b x^{tr}-\b 1 \Big\|_2^2\;+\;\alpha\,R(\b h),
\]
where \(R(\b h)\) is a regularizer (e.g., \(\|\b h\|_2^2\) or a temporal smoothness penalty if filters vary with a parameter), and \(\b 1\) is the all-ones vector providing a unit margin.

Using \(\diag(\b a)\,\b b=\diag(\b b)\,\b a\) for vectors, we rewrite the objective linearly in \(\b h\):
\begin{equation}\label{eq:class_op}
	\min_{\b h}\;\Big\| \diag(\b x^{kn})\,V\,\diag\!\big(V^\top\b x^{tr}\big)\,\Psi\,\b h-\b 1 \Big\|_2^2\;+\;\alpha\,R(\b h),
\end{equation}
so that standard least-squares or ridge regression solvers apply.
The final prediction can then be obtained as
\[
	\b y\;=\;\mathrm{sign}\!\Big( V\,\diag(\Psi\b h)\,V^\top\,\b x^{kn}\Big),
\]
with optional confidence scores given by the pre-sign magnitudes.

\begin{remark}[Multi-class classification]
	For \(C\) classes, construct \(C\) one-vs-all low-pass filters, each trained as above with binary targets.
	At inference, assign the class corresponding to the largest confidence score across the \(C\) filters.
	Calibration  may be applied if probabilistic outputs are needed.
\end{remark}

\subsection{Classification on the inferred DPGs}
Given a static graph \(\mathcal{G}\) and a vertex classification task, we can tune a DPG induced by a \(d\)-dimensional spectral embedding via the threshold \(\delta\) to improve classification accuracy relative to the original topology.
Namely, our idea here can be described as follows: a given static graph \( \mc G \) is not guaranteed to be the best topology for the label propagation in the vertex classification task; instead, we consider a family of DPGs \( \mc G(\delta)\) induced by the original graph that can potentially facilitate better classification.
The original graph should be reachable in this family for specific choices of \( d \) and \( \delta \), \cite{athreya2018statistical}.

Intuitively, the baseline topology \(\mathcal{G}\) may contain heterophilic or noisy connections that hinder label propagation.
The DPG family \(\mathcal{G}(\delta)\) filters edges by latent similarity in the dominant spectral subspace, potentially yielding a graph better aligned with label smoothness.
We select \(\delta\) using a validation split: a portion of known labels is withheld from \(\b x^{kn}\) and used to estimate performance for each candidate \(\delta\).
The best \(\delta\) is then applied to the remaining unlabeled vertices.
In practice, computing the spectral basis \(V(\delta)\) exactly for many \(\delta\) values would be infeasible for large graphs.
However, our interpolation scheme, Algorithm~\ref{alg:interpolationscheme} and Algorithm~\ref{alg:interpolationschemefilter}, provides \(\widetilde{V}(\delta)\) and \(\widetilde{H}(\delta)\) at substantially reduced cost.

\subsection{Results}

\begin{figure}[!t]
	\centering
	\includegraphics[width=0.48\columnwidth]{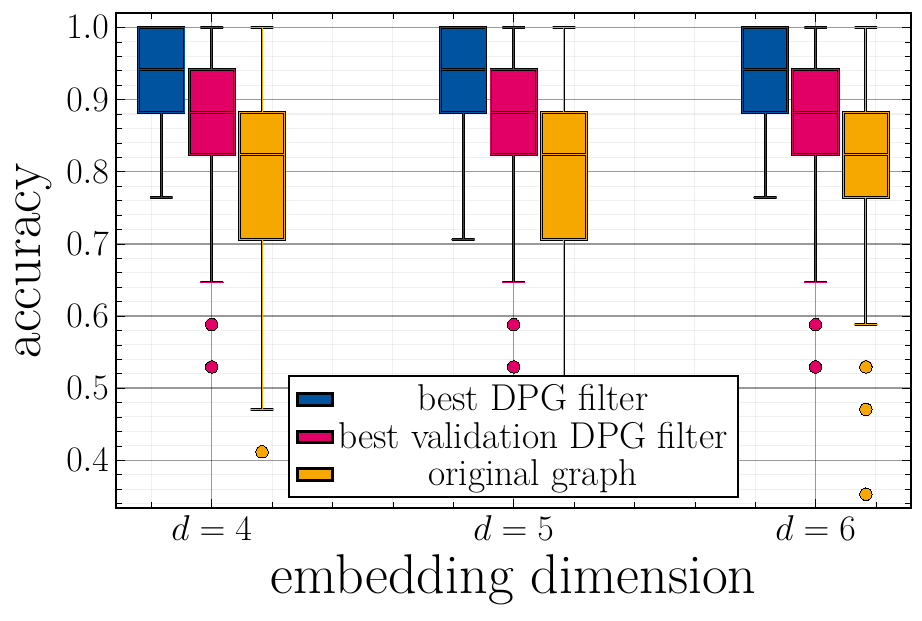}
	\includegraphics[width=0.48\columnwidth]{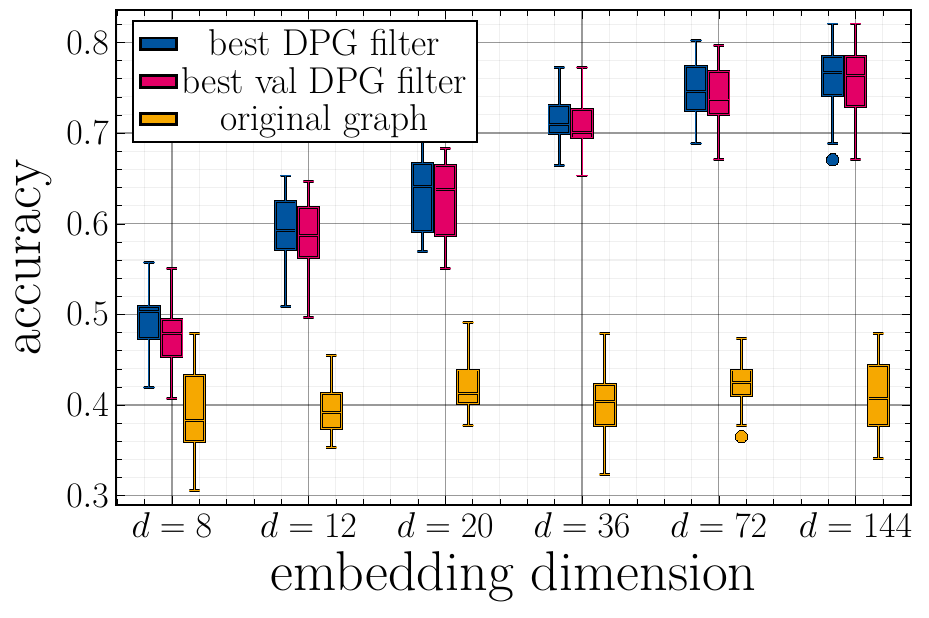}
	\caption{Improvement in the classification accuracy for \texttt{KarateClub} (left) and \texttt{MNIST} similarity graph (right). The low-pass filter retains approximately \(k/n\approx 10\%\) of the spectrum; interpolation uses \(N=10\) Chebyshev nodes. \label{fig:class_res}}
\end{figure}

We evaluate binary vertex classification on two datasets:
\begin{itemize}
	\item \texttt{KarateClub} (\(n=34\), \(78\) edges), unweighted, with a known binary split.
	\item \texttt{MNIST} similarity graph (\(n=1000\)): we sample \(100\) images per digit class. For each image, we connect \(\kappa=8\) Euclidean nearest neighbors, yielding \(\approx 8000\) edges. Edge weights are proportional to Euclidean distance (other choices such as Gaussian kernels are also possible).
\end{itemize}
For each dataset, \(50\%\) of vertex labels are known; \(70\%\) of these serve as training in \eqref{eq:class_op}, \(10\%\) as validation for selecting \(\delta\), and the remaining \(40\%\) are left unlabeled for evaluation. We sweep \(\delta\) to generate the DPG family \(\mathcal{G}(\delta)\) and assess the low-pass filter on each candidate using \(V(\delta)\); for larger graphs, we replace \(V(\delta)\) with its interpolated counterpart \(\widetilde{V}(\delta)\). Figure~\ref{fig:class_res} compares classification accuracy on the original static graph, the best-performing DPG across \(\delta\), and the DPG selected by validation, with distributions obtained via resampling of train/validation/test splits. The results support the proposition: the optimal DPG typically outperforms the static graph, and a validation-driven choice of \(\delta\) closely tracks the best achievable performance within the family.

\begin{remark}[Interpolation for the inferred classification filter]
	Efficient exploration of \(\mathcal{G}(\delta)\) relies on interpolating both the spectral basis \(\widetilde{V}(\delta)\) and the filter \(\widetilde{H}(\delta)\) via Algorithm~\ref{alg:interpolationschemefilter}. One must decide whether the filter taps vector \(\b h(\delta)\) is held fixed, re-estimated for each \(\delta\) (which can be comparable in cost to basis interpolation), or itself interpolated between precomputed instances. In Figure~\ref{fig:class_res}, we re-estimate \(\b h(\delta)\) to obtain the true optimum at each \(\delta\). Alternatively, Lagrange or spline interpolation of \(\b h(\delta)\) is viable if \(\b h(\delta)\) varies smoothly; adding regularization in \eqref{eq:class_op} promotes such smoothness.
\end{remark}

\begin{remark}[Derivative-free selection of \(\delta\)]
	The validation loss \(L(\delta)\) is typically non-smooth: changes in \(\delta\) can alter the graph’s edge set discretely, affecting the spectrum and filter response. Derivative information (e.g., \(\tfrac{d}{d\delta}V(\delta)\)) is often unavailable or unreliable. Derivative-free methods are therefore appropriate: one may perform golden-section search on an interval \([\delta_*,\delta^*]\) if \(L(\delta)\) is approximately unimodal, or use grid search and Bayesian optimization otherwise. Interpolation of \(\widetilde{V}(\delta)\) and \(\widetilde{H}(\delta)\) amortizes the cost across \(\delta\) values.
\end{remark}

% \begin{remark}[Practical considerations]
% - Degree/scale normalization across \(\delta\): rescaling weights to preserve average degree or total weight improves comparability of filters across the DPG family.
% - Choice of \(d\): select via validation, spectral gap heuristics, or information criteria; larger \(d\) captures more structure but increases cost and risk of overfitting.
% - Symmetry and nonnegativity: symmetrize two-sided dot products and apply soft-thresholding to avoid negative weights unless signed models are intended.
% - Large-scale computation: randomized SVD/eigendecomposition and our interpolation scheme make DPG exploration feasible on sizable graphs.
% \end{remark}

Finally, note that the original topology is typically reachable in the DPG family for appropriate \((d,\delta)\) choices (cf. adjacency spectral embedding in latent position graphs \cite{athreya2018statistical}), ensuring that the induced search does not exclude the baseline and allowing principled topology inference tailored to classification.

\section{Conclusion}
\label{sec:conclusion}

In the current work we introduce the notion of low-pass filter interpolation for graph filters using Riemannian interpolation in normal coordinates (Algorithm~\ref{alg:interpolationschemefilter}).
We develop a novel estimate on the interpolation error, Theorem~\ref{thm:int_err}, based on the sensitivity of the Riemannian exponent, Lemma~\ref{lem:sens_exp}
We additionally show that this result characterizes the interpolation error in terms of maximal distance between the base point of the approximation and the subspace trajectory (see Remark~\ref{rem:norm}).
We then outline two application for which the filter interpolation method may be applied: networks with time-dependent features where the evolution of the topology may be obtained through the similarity correction, and an induced dot product representation of a given graph.
The latter induced parametric family is used to formulate a framework where one searches for the optimal network topology \(\mc G(\delta)\) that facilitates better message passing for the vertex classification tasks for both binary and multi-class settings, Figure~\ref{fig:class_res}.

We argue that the suggested optimization of the underlying topology for the message passing scheme can be further utilized for various scenarios, e.g., to inject it into layers of GNNs, apply interpolation for the attention mechanism and for wavelet / or spectral GNNs \cite{xu2019graph, bo2023survey}.
Separately, given  the importance of the spectral information of the graph's shift operator, one can exploit subspace and filter interpolation for the efficient computation of spectral clustering in parametric graph families (e.g., sensor networks with time-dependent sensor coordinates) or in order to modify classical spectral sparsification of networks at the stage of network inference using a threshold parameter to define the underlying graph family.
Finally, we point out that the approaches described in Algorithms~\ref{alg:interpolationscheme} and \ref{alg:interpolationschemefilter} are immediately generalizable to the case of higher-order Laplacian operators for cell complexes and simplicial complexes, which allows the consideration of similar techniques for the processing of signals defined on edges and higher-order structures.

\section*{Acknowledgments}
The authors acknowledge funding by the Deutsche Forschungsgemeinschaft (DFG, German Research Foundation) -- Project number 442047500 through the Collaborative Research Center ``Sparsity and Singular Structures” (SFB 1481). MTS acknowledges funding by the European Union (ERC, HIGH-HOPeS, 101039827). Views and opinions expressed are however those of the author(s) only and do not necessarily reflect those of the European Union or the European Research Council Executive Agency. Neither the European Union nor the granting authority can be held responsible for them.

\bibliographystyle{IEEEtran}
\bibliography{notes}

\vfill

\end{document}